\documentclass{article}

\usepackage{microtype}
\usepackage{graphicx}
\usepackage{booktabs} 

\usepackage{hyperref}

\usepackage[table]{xcolor}
\definecolor{coolbulgarian}{rgb}{0.0, 0.18, 0.39}
\definecolor{maroon}{rgb}{0.76, 0.13, 0.28}

\usepackage[accepted]{icml2024}

\usepackage{amsmath}
\usepackage{amssymb}
\usepackage{mathtools}
\usepackage{amsthm}
\usepackage{bm}

\usepackage[capitalize,noabbrev]{cleveref}

\theoremstyle{plain}

\theoremstyle{definition}

\theoremstyle{remark}

\usepackage[textsize=tiny]{todonotes}

\usepackage{custom}

\icmltitlerunning{Vectorized Conditional Neural Fields}

\begin{document}

\twocolumn[
\icmltitle{Vectorized Conditional Neural Fields: A Framework for Solving Time-dependent Parametric Partial Differential Equations}




\begin{icmlauthorlist}
\icmlauthor{Jan Hagnberger}{mls}
\icmlauthor{Marimuthu Kalimuthu}{mls,simtech,imprs-is}
\icmlauthor{Daniel Musekamp}{mls,imprs-is}
\icmlauthor{Mathias Niepert}{mls,simtech,imprs-is}
\end{icmlauthorlist}

\icmlaffiliation{mls}{Machine Learning and Simulation Lab, Institute for Artificial Intelligence, University of Stuttgart, Stuttgart, Germany}
\icmlaffiliation{simtech}{Stuttgart Center for Simulation Science (SimTech)}
\icmlaffiliation{imprs-is}{International Max Planck Research School for Intelligent Systems (IMPRS-IS)}

\icmlcorrespondingauthor{Jan Hagnberger}{j.hagnberger@gmail.com}

\icmlkeywords{Machine Learning, Partial Differential Equation, PDE, Transformer, Neural Field, ICML}

\vskip 0.3in
]



\printAffiliationsAndNotice{}  

\begin{abstract}
Transformer models are increasingly used for solving Partial Differential Equations (PDEs). Several adaptations have been proposed, all of which suffer from the typical problems of Transformers, such as quadratic memory and time complexity. Furthermore, all prevalent architectures for PDE solving lack at least one of several desirable properties of an ideal surrogate model, such as (i) generalization to PDE parameters not seen during training, (ii) spatial and temporal zero-shot super-resolution, (iii) continuous temporal extrapolation, (iv) support for 1D, 2D, and 3D PDEs, and (v) efficient inference for longer temporal rollouts. To address these limitations, we propose \emph{Vectorized Conditional Neural Fields} (VCNeFs), which represent the solution of time-dependent PDEs as neural fields. Contrary to prior methods, however, VCNeFs compute, for a set of multiple spatio-temporal query points, their solutions in parallel and model their dependencies through attention mechanisms. Moreover, VCNeF can condition the neural field on both the initial conditions and the parameters of the PDEs. An extensive set of experiments demonstrates that VCNeFs are competitive with and often outperform existing ML-based surrogate models.
\end{abstract}

\section{Introduction}
\label{sec:intro}

The simulation of physical systems often involves solving Partial Differential Equations (PDEs), and machine learning-based surrogate models are increasingly used to address this challenging task~\cite{deeponet-lu:2019, graph-neural-operator-li, galerkin-cao}. Utilizing ML for solving PDEs has several advantages, such as faster simulation time than classical numerical PDE solvers, differentiability of the surrogate models \cite{pdebench-takamoto:2022}, and their ability to be used even when the underlying PDEs are not known exactly \cite{fno-li}. However, if knowledge about the PDEs is available, it can be added to the model as in Physics-Informed Neural Networks (PINNs; \citet{pinn-raissi}).

\begin{figure}[t!bp]
    \begin{center}
        \includegraphics[width=1.0\columnwidth]{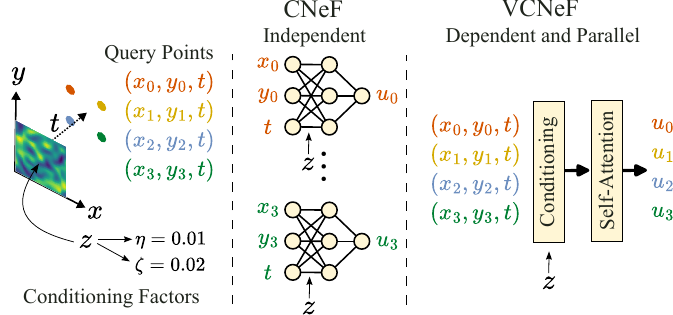}
        \caption{Conditional Neural Field (CNeF) vs proposed Vectorized CNeF (VCNeF) for solving parameterized PDEs.}
        \label{fig:vcnf-teaser}
    \end{center}
\end{figure}

Transformers~\cite{transformers-vaswani} and its numerous variants are successfully used in natural language processing \cite{transfromers-bert-devlin}, speech processing \cite{transformer-conformer-gulati}, and computer vision \cite{transformers-images-dosovitskiy}. Due to their remarkable ability to effectively model long-range dependencies in sequential data and to have favorable scaling behavior, Transformers are used in an increasing number of additional applications.
Transformer models have been gaining traction in Scientific Machine Learning (SciML) to model physical systems~\cite{transformers-physical-systems-geneva:2020}, solve PDEs~\cite{galerkin-cao, oformer-pde-li:2023, scalable-transformer-pde-li:2023, gnot-operator-learn-hao:2023}, and pre-train multiphysics SciML foundation models~\cite{multi-physics-pretrain-avit-mccabe:2023}. 
Meanwhile, recent advances in neural networks for computer graphics tasks have introduced Neural Fields~\cite{neural-fields-xie}, which have proven to be an efficient method to solve PDEs~\cite{siren-inr-sitzmann:2020,crom-pde-yichen-chen:2023,implicit-neural-spatial-repr-pde-chen:2023,dynamics-aware-implicit-neural-repr-dino-yin:2023,operator-learning-neural-fields-general-geometries-serrano:2023}.\blfootnote{The source code of VCNeF framework is available on GitHub: {\href{https://github.com/jhagnberger/vcnef/}{\textcolor{coolbulgarian}{\textbf{https://github.com/jhagnberger/vcnef/}}}}}

Despite these recent advances in neural architectures for PDE solving, current methods lack several of the characteristics of an ideal PDE solver: (i) generalization to different Initial Conditions (ICs), (ii) PDE parameters, (iii) support for 1D, 2D, and 3D PDEs,  (iv) stability over long rollouts, (v) temporal extrapolation, (vi) spatial and temporal super-resolution capabilities, all with affordable cost, high speed, and accuracy. 

Towards developing a model that encompasses these ideal characteristics, we propose \emph{Vectorized Conditional Neural Field} (VCNeF), a linear transformer-based conditional neural field that solves PDEs continuously in time, endowing the model with temporal as well as spatial Zero-Shot Super-Resolution (ZSSR) capabilities. The model introduces a new mechanism to condition a neural field on Initial Conditions (ICs) and PDE parameters to achieve generalization to both ICs and PDE parameter values not seen during training. While modeling the solution using neural fields such as PINNs naturally provide temporal and spatial ZSSR, these methods are inefficient since we need to query them separately for every temporal and spatial location in the domain. We achieve faster training and inference by vectorizing these computations on GPUs. Moreover, the proposed method explicitly models dependencies between multiple simultaneous spatio-temporal queries to the model. 

Concretely, we focus on training and evaluating VCNeF on 1D, 2D, and 3D Initial Value Problems (IVPs) where an IC is given and one predicts multiple future timesteps, as this setting is best suited for real-world applications. The IC could be the data from measurements, and longer rollouts are required to simulate the system under consideration. Additionally, we train our model on multiple PDE parameter values to evaluate its capability to generalize to unseen PDE parameter values.

In summary, we make the following contributions:
\begin{itemize}
    \item A time-continuous transformer-based architecture that represents the solutions to PDEs at any point in time as neural fields, even those not encountered during training, which is accomplished by explicitly incorporating the query time.
    \item We empirically verify that VCNeFs generalize robustly to PDE parameter values not seen during training through effective parameter conditioning while also possessing intrinsic capabilities for spatial and temporal zero-shot superresolution.
    \item A model that naturally provides an implicit vectorization of the spatial coordinates that allows for faster training and inference. It also allows computing the solution of multiple spatial points in one forward pass and exploits spatial dependencies instead of processing them independently.
\end{itemize}

\section{Problem Definition}
In this section, we formally introduce the problem of solving parametric PDEs using neural surrogate models.

\paragraph{Partial Differential Equations.}
Following~\citet{mp-pde-brandstetter}, PDEs over the time dimension, denoted as $t \in [0, T]$, and over multiple spatial dimensions, indicated by $\boldsymbol{x} = (x_x, x_y, x_z, \hdots)^{\top} \in \mathbb{X} \subseteq \mathbb{R}^D$ with $D$ dimensions of a PDE, can be expressed as
\begin{equation}
    \begin{gathered}
        \partial_tu = F(t, \boldsymbol{x}, u, \partial_{\boldsymbol{x}} u, \partial_{\boldsymbol{xx}} u, \hdots) \text{ with } (t, \boldsymbol{x}) \in [0, T] \times \mathbb{X}\\
        u(0, \boldsymbol{x}) = u(0, \cdot) = u^0(\boldsymbol{x}) = u^0 \text{ with } \boldsymbol{x} \in \mathbb{X}\\
        B[u](t, \boldsymbol{x}) = 0 \text{ with } (t, \boldsymbol{x}) \in [0, T] \times \partial \mathbb{X}
    \end{gathered}
    \label{eq:pde}
\end{equation}
where $u: [0, T] \times \mathbb{X} \rightarrow \mathbb{R}^c$ represents the solution function of the PDE that satisfies IC $u(0, \bm{x})$ for time $t = 0$ and the Boundary Conditions (BCs) $B[u](t, \bm{x})$ if $\bm{x}$ is on the boundary $\partial \mathbb{X}$ of the domain $\mathbb{X}$. $c$ denotes the number of output channels or field variables of the PDE. Solving a PDE means determining (an approximation of) the function $u$ that satisfies \cref{eq:pde}. PDEs often contain a parameter, such as the diffusion or viscosity coefficient, which influences their dynamics. We denote the vector of PDE parameter(s) as $\bm{p}$\footnote{Scalars are represented with a small letter (e.g., $a$), vectors with small boldfaced letter (e.g., $\boldsymbol{a}$), and matrices and N-way tensors (s.t. N $\geq$ 3) with a capital boldfaced letter (e.g., $\boldsymbol{A}$).}. The notation $\partial_{\boldsymbol{x}} u, \partial_{\boldsymbol{xx}} u, \hdots$ represents the $i^{th}$ order (where $i \in [1,2,..,n]$) partial derivative $\frac{\partial u}{\partial \boldsymbol{x}}, \frac{\partial^2 u}{\partial \boldsymbol{x}^2}, \hdots, \frac{\partial^n u}{\partial \boldsymbol{x}^n}$. 

\paragraph{Train and Test Data.}
One has to use discretized data generated by a numerical solver to train surrogate models. The temporal domain $[0, T]$ is discretized into $N_t$ timesteps yielding a sequence $(u(t_0, \cdot), u(t_1, \cdot), \hdots, u(t_{N_t-1}, \cdot))$ which describes the evolution of a PDE. $\Delta t = t_{i+1}-t_i$ denotes the temporal step size or resolution. We denote the number of timesteps used for the IC as $N_i$. Since we focus on initial value problems with one timestep as the IC, it holds $N_i = 1$. The spatial domain $\mathbb{X}$ is also transformed into a grid $\boldsymbol{X}$ by discretizing each spatial dimension. Each grid element localizes a point in the spatial domain of the PDE. For 1D PDEs, the grid $\boldsymbol{X} = (\bigl( \boldsymbol{x_i} = (x_{x_i}) \bigr)_{i=1}^{s_x})^{\top} \in \mathbb{R}^{s_x}$ and $s_x$ denotes the spatial resolution (i.e., number of spatial points) of the x-axis. Similarly, for 2D PDEs the grid $\boldsymbol{X} = (\bigl( \boldsymbol{x_i} = (x_{x_i}, x_{y_i}) \bigr)_{i=1}^{s_x \cdot s_y})^{\top} \in \mathbb{R}^{(s_x \cdot s_y) \times 2}$ and $s_x$, $s_y$ denote the spatial resolutions of the x and y axis, respectively. $u(t_i, \boldsymbol{X}) = (u(t_i, \boldsymbol{x_1}), u(t_i, \boldsymbol{x_2}), \hdots, u(t_i, \boldsymbol{x_s}))^{\top} \in \mathbb{R}^{s \times c}$ with $s = s_x \cdot s_y \cdot s_z \cdot \hdots$ contains the solutions at different spatial locations on the grid $\boldsymbol{X}$. The PDE parameters are stacked as a vector $\boldsymbol{p} = (p_1, \hdots, p_j)^{\top} \in \mathbb{R}^j$ where $p_i$ represents the value of a PDE parameter.

A dataset $\mathcal{D} = \{(\boldsymbol{I_1}, \boldsymbol{Y_1}), \hdots, (\boldsymbol{I_N}, \boldsymbol{Y_N})\}$ for each PDE consists of $N$ samples. $\boldsymbol{I_j} = (u(t_0, \boldsymbol{X}), \hdots, u(t_{N_i}, \boldsymbol{X}))$ denotes the solutions given as IC and $\boldsymbol{Y_j} = ((u(t_0, \boldsymbol{X}), \hdots, u(t_{N_t}, \boldsymbol{X}))$ denotes the target sequence of timesteps which represents the trajectory of the PDE.

\paragraph{Training Objective.}
The training objective aims to optimize the parameters $\theta$ (i.e., weights and biases) of the model $f_\theta$ that best approximate the true function $u$ by minimizing the empirical risk over the dataset $\mathcal{D}$
\begin{equation}
    \argmin_{\theta \in \Theta} \sum_{i=1}^N \sum_{j=1}^{N_t} \mathcal{L}\left(f_\theta\left(t_j, \boldsymbol{X} \mid  \boldsymbol{I_i}\right),  \boldsymbol{Y}_{\boldsymbol{i}, j}\right),
    \label{eq:empirical-risk}
\end{equation}
where $\mathcal{L}$ denotes a suitable loss function such as the Mean Squared Error (MSE). $f_\theta(t_j, \boldsymbol{X} |  \boldsymbol{I_i})$ represents the prediction of the neural network for timestep $t_j$ and a grid $\boldsymbol{X}$ given the initial condition $ \boldsymbol{I_i}$.

\section{Background and Preliminaries}
\label{sec:background-preliminaries}

We briefly recall (conditional) neural fields and relate them to solving parametric PDEs.

\paragraph{Neural Fields.}
In physics, a \textit{field} is a quantity that is defined for all spatial and temporal coordinates. Neural Fields (NeFs; \citet{neural-fields-xie}) learn a function $f$ which maps the spatial and temporal coordinates (i.e., $\boldsymbol{x} \in \mathbb{R}^{D}, t \in \mathbb{R}_{+}$ respectively) to a quantity $\boldsymbol{q} \in \mathbb{R}^{c}$. Mathematically, a neural field can be expressed as a function
\begin{equation}
    f_\theta: (\mathbb{R}_+ \times \mathbb{R}^{D}) \rightarrow \mathbb{R}^{c} \text{ with } (t, \boldsymbol{x}) \mapsto \boldsymbol{q}
    \label{eq:neural-field}
\end{equation}
that is parametrized by a neural network with parameters $\theta$. For solving PDEs, the function $f_\theta$ models the solution function $u$, and the quantity $\boldsymbol{q}$ represents the solution's value for the different channels, each representing a physical quantity (e.g., density, velocity, etc.). This architectural design takes inspiration from the Eulerian specification of the flow field from classical field theory, where a field of interest is prescribed both by spatial and temporal coordinates. PINNs \cite{pinn-raissi} are a special case of neural fields with a physics-aware loss function, modeling the solution $u$ as
\begin{equation}
    f_\theta: (\mathbb{R}_{+} \times \mathbb{R}^{D}) \rightarrow \mathbb{R}^{c} \text{ with } (t, \boldsymbol{x}) \mapsto u(t, \boldsymbol{x})
    \label{eq:neural-field-pinn}
\end{equation}
where $f_\theta$ denotes the neural field that maps the input spatial and temporal locations to the solution of the PDE.

\paragraph{Conditional Neural Fields.}
Conditional Neural Fields (CNeFs; \citet{neural-fields-xie}) extend NeFs with a conditioning factor $\boldsymbol{z}$ to influence the output of the neural field. The conditioning factor was originally introduced for computer vision to control the colors or shapes of objects that are being modeled. In contrast, we condition the neural field, which models the solution of the PDE, on the initial value or IC and the PDE parameters (cf. Figure~\ref{fig:vcnf-teaser}). Thus, the conditioning factor influences the entire field.

\section{Method}
\label{sec:method}

In this section, we propose \emph{Vectorized Conditional Neural Fields} by explaining the transition from (conditional) neural fields to vectorized (conditional) neural fields. We also introduce our transformer-based architecture.

\subsection{\emph{Vectorized} Conditional Neural Fields}
Typically, a (conditional) neural field generates the output quantities for all input spatial and temporal coordinates in multiple and independent forward passes. The training and inference times can be improved by processing multiple inputs in parallel on the GPU, which is possible since all forward passes are independent. However, there are spatial dependencies between different input spatial coordinates, particularly for solving PDEs, that will not be exploited with CNeFs or by processing multiple inputs of CNeFs in parallel. Consequently, we propose extending CNeFs to 
\begin{itemize}
    \item take a vector with \emph{arbitrary} spatial coordinates of \emph{variable size} (a set of query points) as input,
    \item exploit the dependencies of the input coordinates when generating the outputs,
    \item generate all outputs for the inputs in one forward pass.
\end{itemize}
Hence, we name our proposed model \emph{Vectorized Conditional Neural Field} since it implicitly generates a vectorization of the input spatial coordinates for a given time $t$. The VCNeF~ model represents a function
\begin{equation}
    \begin{gathered}
        f_\theta: (\mathbb{R}_{+} \times \mathbb{R}^{s \times D}) \rightarrow \mathbb{R}^{s \times c} \\
        \text{ with } (t, \boldsymbol{X}) \mapsto u(t, \boldsymbol{X}) = \begin{pmatrix}
        u(t, \boldsymbol{x_1}) \\
        \vdots \\
        u(t, \boldsymbol{x_s})
        \end{pmatrix}
    \end{gathered}
    \label{eq:vcnef}
\end{equation}
where $u(t, \boldsymbol{x_i})$ denotes the PDE solution for the spatial coordinates $\boldsymbol{x_i}$. Note that we do not impose a structure on the spatial coordinates $\boldsymbol{x_i}$ and that the number of spatial points (i.e., $s$) can be arbitrary. The model can process multiple timesteps $t$ in parallel on the GPU to further improve the training and inference time since VCNeF does not exploit dependencies between the temporal coordinates.

\begin{figure*}[t!hb]
    \begin{center}
        \includegraphics[width=0.925\textwidth]{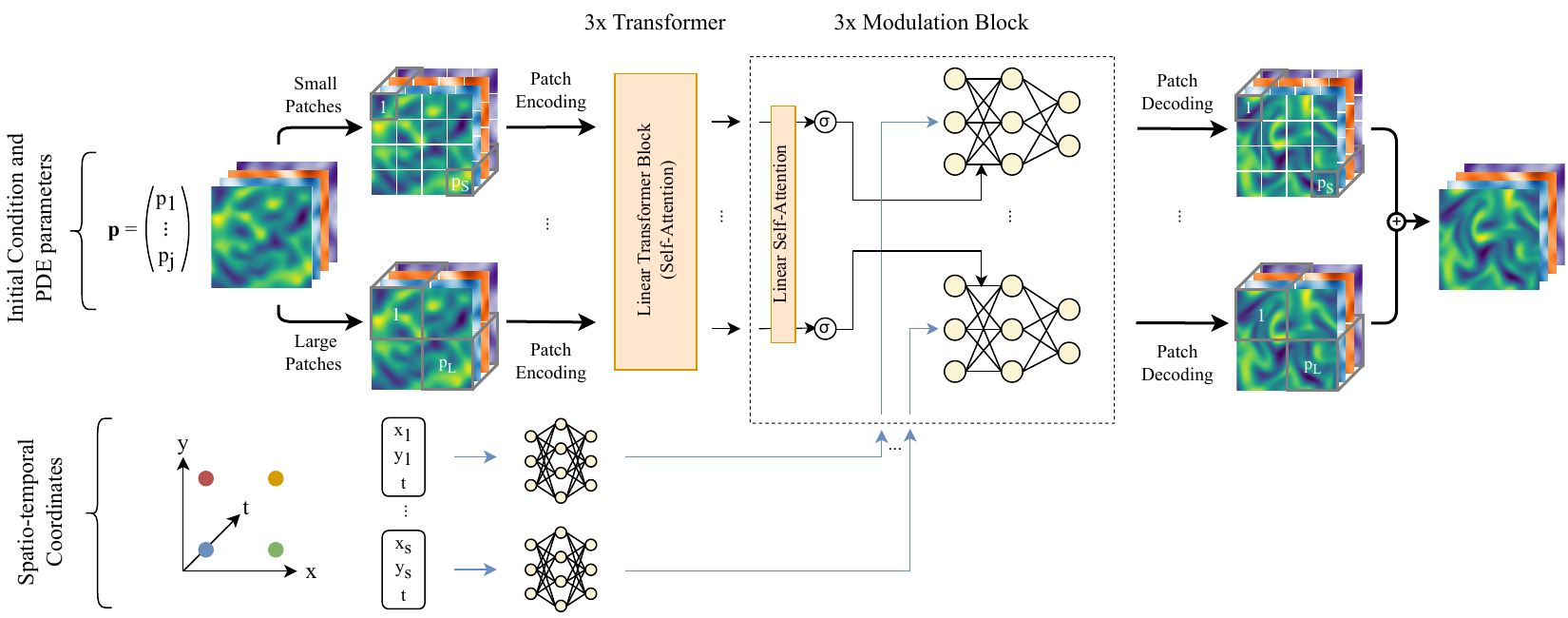}
        \caption{An illustration of the VCNeF architecture for solving parametric time-dependent 2D PDEs. Latent representations of ICs are generated with a multi-scale patching mechanism \cite{multi-scale-vit-chen}. A modulation block consists of self-attention, activation function $\sigma$, and a modulated neural field that uses the scaling of FiLM \cite{film-conditioning-perez:2018} to condition the spatio-temporal coordinates on ICs.}      
        \label{fig:vcnef-architecture}
    \end{center}
\end{figure*}

\subsection{VCNeF for Solving PDEs}
VCNeFs directly learn the solution function $u$ of a PDE by mapping a timestep $t_n$ to a subsequent timestep $t_{n+1}$. Hence, VCNeFs are not autoregressive by design. The model is conditioned on the IC to allow for generalization to different ICs and on the PDE parameters $\boldsymbol{p}$ to generalize to PDE parameter values not seen during training. The VCNeF model can be expressed as a function
\begin{equation}
    \begin{gathered}
        f_\theta: (\mathbb{R}_+ \times \mathbb{R}^{s \times D} \times \mathbb{R}^{s \times c} \times \mathbb{R}^j) \rightarrow \mathbb{R}^{s \times c} \\
        \text{ with } (t,  \boldsymbol{X}, u(0,  \boldsymbol{X}), \boldsymbol{p}) \mapsto u(t,  \boldsymbol{X} | u(0,  \boldsymbol{X}), \boldsymbol{p}),
    \end{gathered}
    \label{eq:vcnef-for-pde}
\end{equation}
where $\theta$ represents the parameters of the neural network, $\boldsymbol{X} \in \mathbb{R}^{s}$ the grid with query spatial coordinates, $t$ the query time, $u(0, \boldsymbol{X})$ the IC, and $\boldsymbol{p}$ the vector of PDE parameters. $u(t, \boldsymbol{X} | u(0, \boldsymbol{X}), \boldsymbol{p})$ denotes the solution function, which depends on the given IC and PDE parameters, that is directly regressed by VCNeF. The shape of the grid $\boldsymbol{X}$ depends on the dimensionality of the PDE.

As a consequence, VCNeFs do not have to generate the PDE trajectory autoregressively. If the complete trajectory is needed, VCNeF can be queried with the desired times $t$ along the trajectory. Furthermore, the model is continuous in time and can do temporal ZSSR (i.e., changing the temporal discretization $\Delta t$ after training) as well as spatial ZSSR (i.e., changing spatial resolution or grid at inference time). Thus, the model can be queried with arbitrary $t \in (0, T]$ and a finer grid $\boldsymbol{\mathcal{X}}$ which is different from the grid $\boldsymbol{X}$ seen during training.
\begin{equation}
    \begin{gathered}
        f_\theta(t, \boldsymbol{\mathcal{X}}, u(0, \boldsymbol{\mathcal{X}}), \boldsymbol{p}) \approx u(t, \boldsymbol{\mathcal{X}} | u(0, \boldsymbol{\mathcal{X}}), \boldsymbol{p}) \quad \forall t \in (0, T]
    \end{gathered}
    \label{eq:vcnef-zssr}
\end{equation}
Alternatively, VCNeFs can be seen as an implementation of a neural operator \cite{neural-operators-kovachki} that is time-continuous and maps the input function (i.e., IC) to an output function that depends on both the IC and time $t$. However, prior work about neural operators \cite{graph-neural-operator-li, fno-li, galerkin-cao, oformer-pde-li:2023} usually does not focus on time continuity. See \cref{app:vcnef-as-no} for more details.

\subsection{Neural Architecture}

We propose a transformer-based VCNeF that applies self-attention to the spatial domain to capture dependencies between the spatial coordinates. The input spatio-temporal coordinates and the physical representation of ICs are represented in a latent space. Both latent representations are fed into modulation blocks that capture spatial dependencies and condition the coordinates on the IC. The output of the modulation blocks, which represent the solution, is then decoded to obtain the representation in the physical space.

\paragraph{Latent Representation of Coordinates.}
The input coordinates, consisting of the query time $t \in \mathbb{R}_+$ that determines the time for which the model's prediction is sought and the spatial coordinates $\boldsymbol{x_i} \in \mathbb{R}^D$, are represented in a latent space. For 1D PDEs, a linear layer is used for encoding, whereas, for 2D and 3D PDEs, the absolute positional encoding (PE; \citet{transformers-vaswani}) to encode time $t$, similar to \citet{ditto-diffusion-temporal-transformer-ovadia:2023} and learnable Fourier features (LFF; \citet{fourier-features-li}) to encode the spatial coordinates are used.
\begin{equation}
    \begin{aligned}
    \text{1D: } \boldsymbol{c_i} &= (t \mathbin\Vert \boldsymbol{x_i})\boldsymbol{W} + \boldsymbol{b} \\
    \text{2D: } \boldsymbol{c_i} &= (\mathtt{PE}(t) \mathbin\Vert \mathtt{LFF}(\boldsymbol{x_i}) \mathbin\Vert \hdots \mathbin\Vert \mathtt{LFF}(\boldsymbol{x_{i+15}}))\boldsymbol{W} + \boldsymbol{b} \\
    \text{3D: } \boldsymbol{c_i} &= (\mathtt{PE}(t) \mathbin\Vert \mathtt{LFF}(\boldsymbol{x_i}) \mathbin\Vert \hdots \mathbin\Vert \mathtt{LFF}(\boldsymbol{x_{i+63}}))\boldsymbol{W} + \boldsymbol{b} \\
    \mathtt{LFF}(\boldsymbol{x}) &= \mathtt{MLP}(\frac{1}{\sqrt{d}}(\cos(\boldsymbol{x}\boldsymbol{W_r}) \mathbin\Vert \sin(\boldsymbol{x}\boldsymbol{W_r}))^\top) \\
    \boldsymbol{C} &= (\boldsymbol{c_1} \mathbin\Vert \hdots \mathbin\Vert \boldsymbol{c_s})^{\top}
    \end{aligned}
    \label{eq:vcnef-latent-coords}
\end{equation}
where $\mathbin\Vert$ stands for the concatenation of two vectors and $\mathtt{MLP}$ denotes a Multi-Layer Perceptron (MLP).

\paragraph{Latent Representation of IC.} 
The input IC is mapped to a latent representation by either applying a shared linear layer to each solution point $u(t, \boldsymbol{x_i})$ or by dividing the spatial domain into non-overlapping patches and applying a linear layer to the patches, akin to Vision Transformers (ViTs; \citet{transformers-images-dosovitskiy}). We divide the spatial domain into patches for 2D and 3D PDEs to reduce the computational costs. However, unlike a traditional ViT, our patch generation has two branches: patches of a smaller size ($p_S = 4$ or $4 \times 4$) and of a larger size ($p_L = 16$ or $16 \times 16$) as proposed in \citet{multi-scale-vit-chen} since we aim to capture the dynamics accurately at multiple scales. 

\begin{equation}
    \begin{aligned}
    \text{1D: } \boldsymbol{z^{(0)}_i} = &(u(t, \boldsymbol{x_i}) \mathbin\Vert \boldsymbol{x_i} \mathbin\Vert \boldsymbol{p})\boldsymbol{W} + \boldsymbol{b} \\
    \text{2D: } \boldsymbol{z^{(0)}_i} = &(u(t, \boldsymbol{x_i}) \mathbin\Vert \hdots \mathbin\Vert u(t, \boldsymbol{x_{i+15}}) \mathbin\Vert \\
    & \boldsymbol{x_i} \mathbin\Vert \hdots \mathbin\Vert \boldsymbol{x_{i+15}} \mathbin\Vert \boldsymbol{p})\boldsymbol{W} + \boldsymbol{b} \\
    \text{3D: } \boldsymbol{z^{(0)}_i} = &(u(t, \boldsymbol{x_i}) \mathbin\Vert \hdots \mathbin\Vert u(t, \boldsymbol{x_{i+63}}) \mathbin\Vert \\
    & \boldsymbol{x_i} \mathbin\Vert \hdots \mathbin\Vert \boldsymbol{x_{i+63}} \mathbin\Vert \boldsymbol{p})\boldsymbol{W} + \boldsymbol{b} \\
    \boldsymbol{Z^{(0)}} = &(\boldsymbol{z^{(0)}_1} \mathbin\Vert \hdots \mathbin\Vert \boldsymbol{z^{(0)}_s})^{\top}
    \end{aligned}
    \label{eq:vcnef-latent-ic}
\end{equation}
A vector in the latent space (i.e., token) either represents the solution on a spatial point (for 1D) or the solution on a patch of spatial points (for 2D and 3D). The grid contains the coordinates where the solutions are sampled in the spatial domain. This information is used when generating the latent representations of the IC to ensure that each latent representation has information about the position. The PDE parameters $\boldsymbol{p}$ are also added to the latent representation. We neglect additional positional encodings to prevent length generalization problems \cite{rope-ruoss} that could prevent changing the spatial resolution after training.

\paragraph{Linear Transformer Encoder for IC.}
We utilize a Linear Transformer \cite{linear-transformers-katharopoulos:2020} with self-attention in our VCNeF architecture to generate an attention-refined latent representation of the IC $\boldsymbol{Z^{(0)}}$. The global receptive field of the Transformer allows the proposed architecture to capture global spatial dependencies in the IC, although each token contains only local spatial information. Intuitively, the Transformer outputs latent representations that incorporate the entire spatial solution and not only a single spatial point or a subset of spatial points. We assume that this is beneficial to generate a better representation of the IC to condition the input coordinates accordingly.
\begin{equation}
    \bm{Z^{(n+1)}} = \mathtt{Transformer\_Block}\left(\bm{Z^{(n)}}\right)
    \label{eq:transformer-encoder}
\end{equation}
where $\mathtt{Transformer\_Block}(\cdot)$ is a Linear Transformer block with self-attention and $n$ denotes the $n^{th}$  block.

\paragraph{Modulation of Coordinates based on IC.}
The modulation blocks condition the input coordinates on the input IC $\boldsymbol{Z^{(3)}}$ by modulating the latent representation $\boldsymbol{C}$ of the coordinates. The block contains self-attention, a non-linearity $\sigma$, a modulation mechanism similar to Feature-wise Linear Modulation (FiLM; \citet{film-conditioning-perez:2018}), layer normalization, residual connections, and an MLP. However, the conditioning mechanism uses only the scaling (i.e., pointwise multiplication) of FiLM and omits the shift (i.e., pointwise addition). A modulation block is expressed as
\begin{equation}
    \begin{aligned}
        \bm{Z^{(m+1)}} &= \mathtt{Modulation\_Block}\left(\bm{C}, \bm{Z^{(m)}}\right) \\
        &= \mathtt{MLP}\left(\sigma\left(\mathtt{Self\_Attn}\left(\bm{Z^{(m)}}\right)\right) \circ \mathtt{MLP}(\bm{C})\right) \\
        \sigma(\bm{X}) &= \mathtt{ELU}(\bm{X}) + 1
    \end{aligned}
    \label{eq:vcnef-modulation-block}
\end{equation}

where $\boldsymbol{Z^{(3)}} \in \mathbb{R}^{s \times d}$ represents the IC, $\boldsymbol{C} \in \mathbb{R}^{s \times d}$ denotes the latent representation of the input coordinates, $\circ$ represents the Hadamard product, and $m$ is the $m^{th}$ modulation block. The residual connections and layer normalization are omitted in \cref{eq:vcnef-modulation-block} for the sake of simplicity. The modulation blocks condition the spatio-temporal coordinates on the IC and PDE parameter values, and spatial self-attention incorporates dependencies between the queried spatial coordinates.

\paragraph{Decoding the Solution's Latent Representation.}
The solution's latent representation $\boldsymbol{Z^{(6)}}$ is mapped back to the physical space by either applying an MLP for 1D or by mapping the latent representations to small and large patches and outputting the weighted sum of the small and large patches for 2D and 3D.

\section{Properties of VCNeFs}

The proposed VCNeF model has the following properties.

\paragraph{Spatial and Temporal ZSSR.}
VCNeF can be trained on lower spatial and temporal resolutions and used for high-resolution spatial and temporal inference since the model is space and time continuous. To do so, the model can be queried with finer coordinates (i.e., intermediate spatial and temporal coordinates). Training on low-resolution data requires less computational resources and saves computing time, while inference at high-resolution data minimizes the risk of missing crucial dynamics.

\paragraph{Accelerated Training and Inference.}
The training and inference of VCNeF are accelerated by processing multiple temporal coordinates in parallel on the GPU. If the solution of multiple timesteps (e.g., $t \in \{t_1, t_2, \hdots, t_{N_t} \}$) is to be predicted, VCNeF can calculate the solution of the timesteps in parallel due to the fact that the predictions of $u(t, \cdot)$ are independent of each other. The proposed architecture uses linear attention. Nonetheless, linear attention can be replaced with an arbitrary attention mechanism. The runtime and memory consumption are influenced by the spatial resolution $s = s_x \cdot s_y \cdot s_z \cdot \hdots$ and the cardinality $N_t$ of the queried timesteps. For 1D, the model has a time and space complexity of $\mathcal{O}\left(s_x \cdot N_t\right)$. For 2D, the complexity is of $\mathcal{O}\left(\left(\frac{s_x \cdot s_y}{{p_S}^2} + \frac{s_x \cdot s_y}{{p_L}^2}\right) \cdot N_t\right)$ where $s_x$ and $s_y$ denote the spatial resolution of the x and y-axis, respectively, and $p_S, p_L$ denote the patch sizes. We omit encoding and decoding, which include the channels $c$, for the sake of simplicity.

\paragraph{Physics-Informed VCNeF.}
The loss function of VCNeF can be easily extended with a physics-informed loss as in PINNs \cite{pinn-raissi} since VCNeF directly models the solution function $u$ and therefore, the derivatives can be computed with automatic differentiation \cite{hips-autograd-maclaurin:2015,autodiff-pytorch-paszke:2017}.

\paragraph{Randomized Starting Points Training.}
We suggest conditioning the model not only on the IC but also on randomly sampled timesteps along the trajectory in the training phase as data augmentation to improve the model's performance further.

\section{Related Work}

\paragraph{Physics-Informed Neural Networks and Neural Operators.}
A common approach for solving PDEs is Physics-Informed Neural Networks \cite{pinn-raissi} that model the underlying solution function $u$. Although PINNs are space-and-time continuous within the specified domain, they are finite-dimensional and hence cannot perform temporal extrapolation. Additionally, PINNs generate the solution for all input spatial coordinates independently without further exploitation of structural dependencies (\cref{fig:vcnf-teaser}). A PDE-specific loss function allows the model to learn the underlying solution function, which satisfies the PDE equation. However, PINNs can still fail to approximate the PDE solutions because of complex loss landscapes~\cite{pinn-failure-modes-krishnapriyan:2021}. In contrast, neural operators \citep{graph-neural-operator-li} learn a mapping between two infinite-dimensional spaces or two functions where the function represents the solution function of the PDE. Consequently, neural operators are theoretically continuous in space and time, but current implementations usually have limited support for being continuous in space and time. Furthermore, neural operators generate the solution for all spatial coordinates in a single forward pass~\citep{deeponet-lu:2019} and leverage the spatial dependencies of the solution by processing the spatial coordinates in one forward pass. The Fourier Neural Operator (FNO; \citet{fno-li}) is a prevalent instantiation of a neural operator that is based on Fourier transforms. Physics-informed neural operators \cite{physics-neural-operator-li} extend neural operators with a PDE-specific loss to further improve the accuracy of the model. VCNeFs can be seen as a combination of PINNs and neural operators since VCNeFs can be queried continuously over time like PINNs but process the spatial coordinates in a single forward pass, exploiting the spatial dependencies between the queried coordinates as existing neural operator implementations do.

\begin{table}[t!]
    \centering
    \scalebox{0.85}{
        \begin{tabular}{ cccc }
            \toprule
            PDE & \makecell{Timesteps} & \makecell{Spatial res.} & \makecell{PDE parameters} \\
            \midrule
            1D Burgers & 41 & 256 & $\nu$ = 0.001 \\
            1D Advection & 41 & 256 & $\beta$ = 0.1 \\
            1D CNS & 41 & 256 & $\eta$ = $\zeta$ = 0.007 \\
            \midrule
            2D CNS & 21 & $64 \times 64$ & $\eta$ = $\zeta$ = 0.01 \\
            \midrule
            3D CNS & 11 & $32 \times 32 \times 32$ & $\eta$ = $\zeta = 10^{-8}$ \\
            \bottomrule
        \end{tabular}
    }
    \caption{Fixed PDE parameters used in our experiments.}
    \label{tab:pde_params_experiments_single}
\end{table}

\paragraph{Transformers for Solving PDEs.}
Transformers are increasingly being utilized for modeling physical systems or PDEs. Previous works can be divided into using Transformers for applying temporal self-attention \cite{transformers-physical-systems-geneva:2020} to model the temporal dependencies or applying spatial self-attention for capturing spatial dependencies of the PDE \cite{galerkin-cao, oformer-pde-li:2023, scalable-transformer-pde-li:2023}. Applying the spatial self-attention as in Fourier and Galerkin Transformer \cite{galerkin-cao} or in OFormer \cite{oformer-pde-li:2023} yields a neural operator \cite{neural-operators-kovachki} endowing the model with spatial ZSSR capabilities. Since these models do not consider time as an additional input, they are not time-continuous and, hence, fixed to a trained temporal discretization. To remedy the issue, the diffusion-inspired temporal Transformer operator \cite{ditto-diffusion-temporal-transformer-ovadia:2023} uses the time to condition the input solution, thereby supporting a flexible temporal discretization for inference. VCNeFs use spatial self-attention similar to the Fourier and Galerkin Transformers as well as the OFormer. However, our architecture is time-continuous by employing a conditional neural field that modulates spatio-temporal coordinates based on ICs and PDE parameters. DiTTO uses a UNet architecture that is enhanced with self-attention and conditioned on time by modulating the activations with scaling. In contrast, our architecture mainly relies on a transformer architecture with spatial (linear) self-attention and a modulated neural field.

\paragraph{Solving Parametric PDEs.}
Although ML-based methods have shown great success in solving PDEs, they often do not consider PDE parameters as input, resulting in failures to generalize to unseen parameter values. Recent works such as CAPE \cite{cape-takamoto:2023}, PDERefiner \cite{pde-refiner-long-term-rollouts-neural-pde-lippe:2023}, and MP-PDE \cite{mp-pde-brandstetter} consider the PDE parameters as additional model input. Along the same lines, our proposed model also considers the PDE parameter to improve the generalization error of unseen PDE parameter values.

\begin{table}[t!]
    \begin{center}
        \scalebox{0.75}{
            \begin{tabular}{ llll }
                \toprule
                PDE & Model & nRMSE (\textbf{$\downarrow$}) & bRMSE (\textbf{$\downarrow$}) \\
                \midrule
                \multirow{7}{*}{Burgers} & FNO & 0.0987 & 0.0225 \\
                & MP-PDE & 0.3046 (+208.7\%) & 0.0725 (+221.7\%) \\
                & UNet & \textbf{0.0566} (-42.6\%) & 0.0259 (+14.7\%) \\
                & CORAL & 0.2221 (+125.1\%) & 0.0515 (+128.2\%) \\
                & Galerkin & 0.1651 (+67.3\%) & 0.0366 (+62.3\%) \\
                & OFormer & 0.1035 (+4.9\%) & \underline{0.0215} (-4.5\%) \\
                & VCNeF & 0.0824 (-16.5\%) & 0.0228 (+1.3\%) \\
                & VCNeF-R & \underline{0.0784} (-20.6\%) & \textbf{0.0179} (-20.8\%) \\
                \midrule
                \multirow{7}{*}{Advection} & FNO & 0.0190 & 0.0239  \\
                & MP-PDE & 0.0195 (+2.7\%) & 0.0283 (+18.4\%) \\
                & UNet & \textbf{0.0079} (-58.4\%) & 0.0129 (-45.9\%) \\
                & CORAL & 0.0198 (+4.3\%) & 0.0127 (-46.8\%) \\
                & Galerkin & 0.0621 (+227.1\%) & 0.0349 (+46.2\%) \\
                & OFormer & 0.0118 (-38.0\%) & \underline{0.0073} (-69.6\%) \\
                & VCNeF & 0.0165 (-13.0\%) & 0.0088 (-63.2\%) \\
                & VCNeF-R & \underline{0.0113} (-40.5\%) & \textbf{0.0040} (-83.3\%) \\
                \midrule
                \multirow{6}{*}{1D CNS} & FNO & 0.5722 & 1.9797 \\
                & UNet & \underline{0.2270} (-60.3\%) & \textbf{1.0399} (-47.5\%) \\
                & CORAL & 0.5993 (+4.7\%) & 1.5908 (-19.6\%) \\
                & Galerkin & 0.7019 (+22.7\%) & 3.0143 (+52.3\%) \\
                & OFormer & 0.4415 (-22.9\%) & 2.0478 (+3.4\%) \\
                & VCNeF & 0.2943 (-48.6\%) & 1.3496 (-31.8\%) \\
                & VCNeF-R & \textbf{0.2029} (-64.5\%) & \underline{1.1366} (-42.6\%) \\
                \midrule
                \multirow{4}{*}{2D CNS} & FNO & \underline{0.5625} & \underline{0.2332} \\
                & UNet & 1.4240 (+153.2\%) & 0.3703 (+58.8\%) \\
                & Galerkin & 0.6702 (+19.2\%) & 0.8219 (+252.4\%) \\
                & VCNeF & \textbf{0.1994} (-64.6\%) & \textbf{0.0904} (-61.2\%) \\
                \midrule
                \multirow{2}{*}{3D CNS} & FNO & 0.8138 & 6.0407 \\
                & VCNeF & \textbf{0.7086} (-12.9\%) & \textbf{4.8922} (-19.0\%) \\
                \bottomrule
            \end{tabular}
        }
        \caption{Errors of surrogate models trained and tested on the same spatial and temporal resolution with a fixed PDE parameter. nRMSE and bRMSE denote the normalized and RMSE at the boundaries, respectively. Values in parentheses indicate the percentage deviation to the FNO as a strong baseline in terms of accuracy, memory consumption, and runtime. Underlined values indicate the second-best errors.}
        \label{tab:errors-generalization-ic}
    \end{center}
\end{table}

\paragraph{Implicit Neural Representations (INR).} 
Neural Fields (NeFs) has become widely popular in signal processing~\cite{siren-inr-sitzmann:2020}, computer vision~\cite{occupancy-networks-inr-mescheder:2019}, computer graphics~\cite{physics-informed-neural-fields-chu:2022}, and recently in SciML for solving PDEs~\cite{crom-pde-yichen-chen:2023,dynamics-aware-implicit-neural-repr-dino-yin:2023,implicit-neural-spatial-repr-pde-chen:2023,operator-learning-neural-fields-general-geometries-serrano:2023}.
The prevalent INR models for PDE solving follow the ``Encode-Process-Decode'' paradigm. DiNO~\cite{dynamics-aware-implicit-neural-repr-dino-yin:2023} has an encoder, a Neural ODE (NODE) to model dynamics, and a decoder. CORAL, an improvement to DiNO, has a two-step training procedure whereby the input and output INR modules with shared parameters are trained first, and subsequently, the dynamics modeling block is trained using the learned latent codes. DiNO and CORAL utilize an INR to encode and decode the PDE solution in a latent space and a NODE to propagate the dynamics in latent space, while our approach utilizes a neural field to represent the entire PDE solution encompassing both the spatial and temporal dependencies within a shared space.

\section{Experiments}
\label{sec:experiments}

We aim to answer the following research questions.

\begin{description}
    \item[Q1:] How effective are VCNeFs compared to the state-of-the-art (SOTA) methods when trained and tested for the same PDE parameter value?
    \item[Q2:] How well can VCNeFs generalize to PDE parameter values not seen during training?
    \item[Q3:] How well can VCNeFs do spatial and temporal zero-shot super-resolution?
    \item[Q4:] Does training on initial conditions sampled from training trajectories improve the accuracy?
    \item[Q5:] Does the vectorization provide a speed-up, and what is the model's scaling behavior?
\end{description}

\begin{figure}[t!]
    \begin{center}
        \includegraphics[width=0.9\columnwidth]{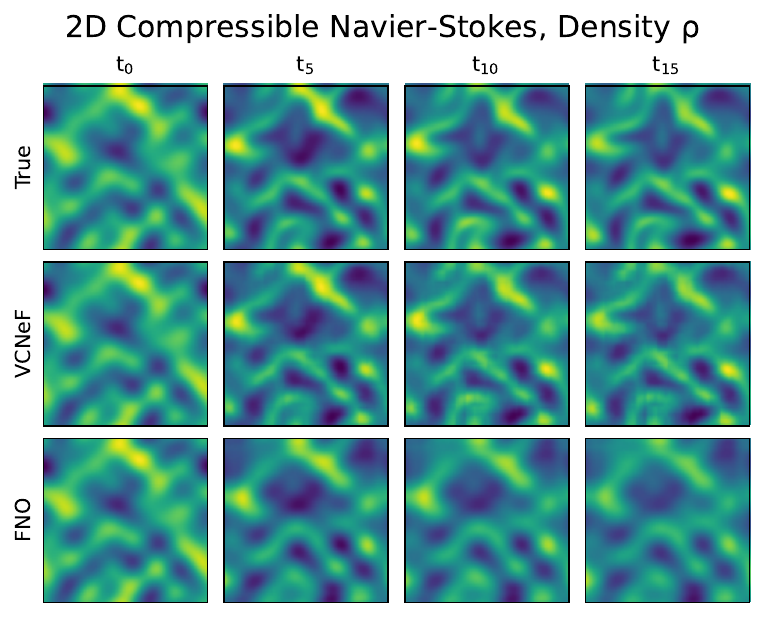}
        \caption{Example predictions for density of 2D CNS.}
        \label{fig:density-2d-cfd-vcnf-fno}
    \end{center}
\end{figure}

\begin{figure}[t!]
    \begin{center}
        \includegraphics[width=1.0\columnwidth]{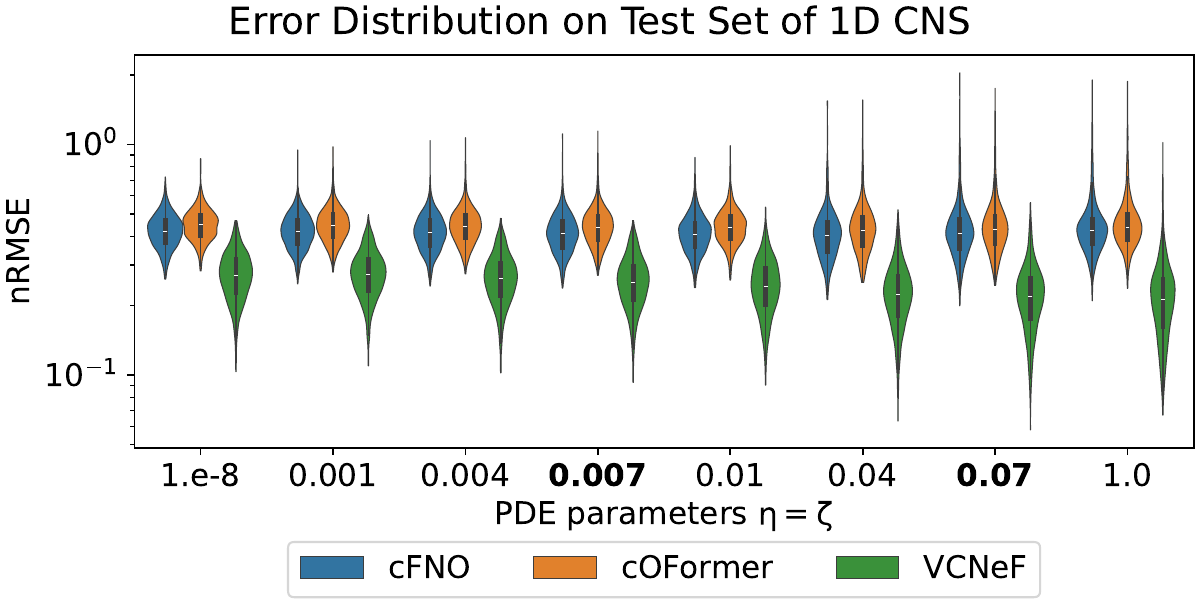}
        \caption{Error distribution of samples in the test set of 1D CNS. Boldfaced are the unseen PDE parameter values.}
        \label{fig:violin-multi-cfd}
    \end{center}
\end{figure}

\subsection{Datasets}

We conduct experiments on the following hydrodynamical equations of parametric PDE datasets from PDEBench~\cite{pdebench-takamoto:2022}: 1D \textbf{Burgers'}, 1D \textbf{Advection}, and 1D, 2D, and 3D \textbf{Compressible Navier-Stokes} (CNS).

\begin{table}[!t]
    \center
    \scalebox{0.75}{
        \begin{tabular}{ cclll }
            \toprule
            PDE & Spatial res. & Model & nRMSE (\textbf{$\downarrow$}) & bRMSE (\textbf{$\downarrow$}) \\
            \midrule
            \multirow{9}{*}{Burgers} & \colored & \colored FNO & \colored \underline{0.0987} & \colored \underline{0.0225} \\
            & \colored & \colored OFormer & \colored 0.1035 & \colored \textbf{0.0215} \\
            & \multirow{-3}{*}{\colored 256} & \colored VCNeF & \colored \textbf{0.0824} & \colored 0.0228 \\
            \cmidrule{2-5}
            & \multirow{3}{*}{512} & FNO & 0.2557 & 0.0566 \\
            & & OFormer & \underline{0.1092} & \textbf{0.0228} \\
            & & VCNeF & \textbf{0.0832} & \underline{0.0229} \\
            \cmidrule{2-5}
            & \multirow{3}{*}{1024} & FNO & 0.3488 & 0.0766 \\
            & & OFormer & \underline{0.1102} & \underline{0.0233} \\
            & & VCNeF & \textbf{0.0839} & \textbf{0.0230} \\
            \midrule
            \multirow{9}{*}{1D CNS} & \colored & \colored FNO & \colored 0.5722 & \colored \underline{1.9797} \\
            & \colored & \colored OFormer & \colored \underline{0.4415} & \colored 2.0478 \\
            & \colored \multirow{-3}{*}{256} & \colored VCNeF & \colored \textbf{0.2943} & \colored \textbf{1.3496} \\
            \cmidrule{2-5}
            & \multirow{3}{*}{512} & FNO & 0.6610 & 2.7683 \\
            & & OFormer & \underline{0.4657} & \underline{2.5618} \\
            & & VCNeF & \textbf{0.2943} & \textbf{1.3502} \\
            \cmidrule{2-5}
            & \multirow{3}{*}{1024} & FNO & 0.7320 & 3.5258 \\
            & & OFormer & \underline{0.4655} & \underline{2.5526} \\
            & & VCNeF & \textbf{0.2943} & \textbf{1.3510} \\
            \midrule
            \multirow{4}{*}{2D CNS} & \colored & \colored FNO & \colored 0.5625 & \colored 0.2332 \\
            & \multirow{-2}{*}{\colored $64 \times 64$} & \colored VCNeF & \colored \textbf{0.1994} & \colored \textbf{0.0904} \\
            \cmidrule{2-5}
            & \multirow{2}{*}{$128 \times 128$} & FNO & 0.8693 & 2.3944 \\
            & & VCNeF & \textbf{0.4016} & \textbf{0.2280} \\
            \midrule
            \multirow{6}{*}{3D CNS} & \colored & \colored FNO & \colored 0.8138 & \colored 6.0407 \\
            & \multirow{-2}{*}{\colored $32 \times 32 \times 32$} & \colored VCNeF & \colored \textbf{0.7086} & \colored \textbf{4.8922} \\
            \cmidrule{2-5}
            & \multirow{2}{*}{$64 \times 64 \times 64$} & FNO & 0.9452 & 8.7068 \\
            & & VCNeF & \textbf{0.7228} & \textbf{5.1495} \\
            \cmidrule{2-5}
            & \multirow{2}{*}{$128 \times 128 \times 128$} & FNO & 1.0077 & 9.8633 \\
            & & VCNeF & \textbf{0.7270} & \textbf{5.3208} \\
            \bottomrule
        \end{tabular}
    }
    \caption{Normalized RMSEs and RMSEs at the boundary for the spatial ZSSR experiments. The models are trained on the spatial resolutions given in the grey columns and are only tested on the additional spatial resolutions. The temporal resolution is the same as during training.
    \label{tab:spatial-zssr}}
\end{table}

\subsection{Setup and Baselines}

We train and test the models with a single timestep as an initial condition and predict multiple future steps. This setting is well-motivated by applications that need solutions to initial value problems. We use the PDE parameter values in Table \ref{tab:pde_params_experiments_single} for the experiments where we train and test with one fixed PDE parameter value. Additionally, we conduct experiments on multiple PDE parameter values, including values not seen during training, to test the models' generalization capabilities.

We choose FNO \cite{fno-li}, MP-PDE~\cite{mp-pde-brandstetter}, UNet from PDEArena~\cite{generalized-pde-modeling-gupta:2023}, CORAL~\cite{operator-learning-neural-fields-general-geometries-serrano:2023}, Galerkin Transformer \cite{galerkin-cao}, and OFormer \cite{oformer-pde-li:2023} as baselines. The predictions of FNO, UNet, OFormer, and Galerkin Transformer are achieved in an autoregressive fashion, while VCNeF predicts the entire trajectory of the simulation directly in one forward pass or single shot.

The reported numbers are the mean values of two training runs with different initializations, and the full results are in \cref{app:results}.

\begin{table}[t!]
    \center
    \scalebox{0.75}{
        \begin{tabular}{ cclll }
            \toprule
            PDE & Temporal res. & Model & nRMSE (\textbf{$\downarrow$}) & bRMSE (\textbf{$\downarrow$})\\
            \midrule
            \multirow{9}{*}{Burgers} & \colored & \colored FNO & \colored \underline{0.0987} & \colored \textbf{0.0225} \\
            & \colored & \colored CORAL & \colored 0.2221 & \colored 0.0515 \\
            & \multirow{-3}{*}{\colored 41} & \colored VCNeF & \colored \textbf{0.0824} & \colored \underline{0.0228} \\
            \cmidrule{2-5}
            & \multirow{3}{*}{101} & FNO + Interp. & \underline{0.1116} & \underline{0.0279} \\
            & & CORAL & 0.5298 & 0.1682 \\
            & & VCNeF & \textbf{0.0829} & \textbf{0.0234} \\
            \cmidrule{2-5}
            & \multirow{3}{*}{201} & FNO + Interp. & \underline{0.1154} & \underline{0.0294} \\
            & & CORAL & 0.6186 & 0.2013 \\
            & & VCNeF & \textbf{0.0831} & \textbf{0.0236} \\
            \midrule
            \multirow{9}{*}{Advection} & \colored & \colored FNO & \colored \underline{0.0190} & \colored 0.0239 \\
            & \colored & \colored CORAL & \colored 0.0198 & \colored \underline{0.0127} \\
            & \multirow{-3}{*}{\colored 41} & \colored VCNeF & \colored \textbf{0.0165} & \colored \textbf{0.0088} \\
            \cmidrule{2-5}
            & \multirow{3}{*}{101} & FNO + Interp. & \underline{0.0234} & \underline{0.0242} \\
            & & CORAL & 0.8970 & 0.4770 \\
            & & VCNeF & \textbf{0.0165} & \textbf{0.0088} \\
            \cmidrule{2-5}
            & \multirow{3}{*}{201} & FNO + Interp. & \underline{0.0258} & \underline{0.0247} \\
            & & CORAL & 0.9656 & 0.5376 \\
            & & VCNeF & \textbf{0.0165} & \textbf{0.0088} \\
            \midrule
            \multirow{6}{*}{1D CNS} & \colored & \colored FNO & \colored \underline{0.5722} & \colored \underline{1.9797} \\
            & \colored & \colored CORAL & \colored 0.5993 & \colored 1.5908 \\
            & \multirow{-3}{*}{\colored 41} & \colored VCNeF & \colored \textbf{0.2943} & \colored \textbf{1.3496} \\
            \cmidrule{2-5}
            & \multirow{3}{*}{82} & FNO + Interp. & \underline{0.5667} & \underline{1.9639} \\
            & & CORAL & 1.1524 & 3.7960 \\
            & & VCNeF & \textbf{0.2965} & \textbf{1.3741} \\
            \midrule
            \multirow{4}{*}{3D CNS} & \colored & \colored FNO & \colored 0.8138 & \colored 6.0407 \\
            & \multirow{-2}{*}{\colored 11} & \colored VCNeF & \colored \textbf{0.7086} & \colored \textbf{4.8922} \\
            \cmidrule{2-5}
            & \multirow{2}{*}{21} & FNO + Interp. & 0.8099 & 6.1938 \\
            & & VCNeF & \textbf{0.7106} & \textbf{5.1446} \\
            \bottomrule
        \end{tabular}
    }
    \caption{Error values for temporal ZSSR. The models are trained on the temporal resolution given in the grey columns and additionally tested on higher temporal resolutions. The spatial resolution is $s = 256$ for 1D and $s = 32 \times 32 \times 32$ for 3D (same as during training).
    \label{tab:temporal-zssr}}
\end{table}

\subsection{Results}

\paragraph{Q1.}
We test VCNeF's generalization ability to different ICs by evaluating the models on the corresponding test sets of PDEBench. \cref{tab:errors-generalization-ic} shows the errors of the baseline models trained and tested on the selected PDEs and \cref{fig:density-2d-cfd-vcnf-fno} an example prediction for 2D CNS. The results demonstrate that our model performs competitively with SOTA methods for solving PDEs.

\paragraph{Q2.}
To evaluate the performance and effectiveness of VCNeF and its PDE parameter conditioning, we train VCNeF, a PDE parameter conditioned FNO (cFNO; \citet{cape-takamoto:2023}), and OFormer that has been modified to accept PDE parameter as additional input\footnote{We encode the parameter values as an additional channel.} on a set of PDE parameter values and test them on another unseen set of parameter values. \cref{fig:violin-multi-cfd} shows the error distribution for 1D CNS. We observe that VCNeF generalizes better to unseen PDE parameters than the other baseline models. See~\cref{app:pde-param-generalization-results} for results on other 1D PDEs (Burgers' and Advection).

\begin{figure}[!t]
    \centering
    \includegraphics[width=1.0\columnwidth]{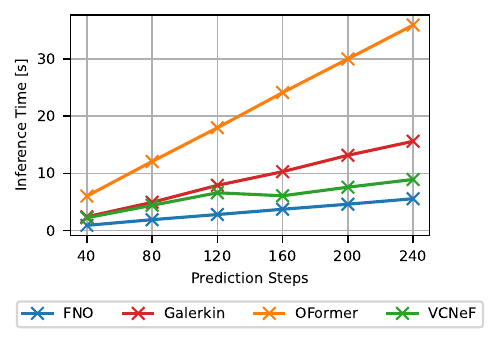}
    \caption{Inference times of models trained on Burgers with $s = 256$, predicting different numbers of timesteps in the future.}
    \label{tab:inference-times}
\end{figure}

\paragraph{Q3.}
We also evaluate the model's ability to do spatial and temporal ZSSR by training the model with a reduced spatial and temporal resolution and testing for a higher resolution. The results in \cref{tab:spatial-zssr} show that the OFormer and VCNeF seem to have spatial ZSSR capabilities since there is no significant increase in error even when the spatial resolution is four times the training resolution. For the case of FNO on 1D Burgers, we observe a near multiplicative increase of error as the factor of resolution increases (2x and 4x).

Temporal ZSSR means that the model predicts a trajectory with a smaller temporal step size $\Delta t$ than encountered during training (i.e., the trajectory is divided into more timesteps). This is possible due to VCNeF's support for continuous-time inference. We use FNO with linear interpolation between the time steps as a baseline since we train FNO in an autoregressive fashion, which fixes the temporal discretization of the trained model. \cref{tab:temporal-zssr} shows a negligible increase in error for VCNeF, meaning that the model has temporal ZSSR capabilities and learns the dependency between solution and time. Doing interpolation seems to be only effective for smooth targets such as CNS.

\paragraph{Q4.}
Conditioning VCNeF on random starting points sampled in the temporal extent of the trajectory $u(t,  \boldsymbol{X})$ with $t \sim \mathcal{U}\{0, T\}$ in addition to the IC $u(0,  \boldsymbol{X})$ during training further enhances the performance. \cref{tab:errors-generalization-ic} shows the performance improvement of the randomized VCNeF (VCNeF-R) over the non-randomized training of VCNeF. We attribute the boost in the performance of VCNeF-R to the diversity of encountered ICs that the model processes during training.

\paragraph{Q5.}
We also measure the inference times of the proposed and baseline models. The times are the raw inference times without measuring the time needed to transfer the data from the host device to the GPU on a single NVIDIA A100-SXM4 80GB GPU. We measure the time on the test set of 1D Burgers (i.e., 1k ICs, predicting 40 to 240 timesteps in the future) with a batch size of 64 and a spatial resolution of $s = 256$. The times in \cref{tab:inference-times} demonstrate that VCNeF is significantly faster than the other transformer-based counterparts. However, the speed-up is traded for a higher GPU memory consumption as shown in \cref{tab:inference-times-verbose}. To mitigate the high GPU memory consumption, the VCNeF can also be used to generate the solutions sequentially, drastically reducing the GPU memory consumption but resulting in a higher inference time for the trajectory. Thus, the proposed model allows for a trade-off between the inference time and GPU memory consumption when generating the solutions~(see \cref{app:inference-times}).

\section{Limitations}
The limitations are two-fold. (i) Using VCNeF to generate trajectories consisting of multiple timesteps in parallel is limited to GPUs with larger memory and hence incurs higher costs. Nevertheless, the sequential VCNeF does not have this limitation (see \cref{app:inference-times}). (ii) VCNeF utilizes 2D and 3D convolutional layers for 2D and 3D PDEs to partition the spatial domain into (non-overlapping) patches, which restricts the model to regular grids. However, the VCNeF model can be used with a different encoder, such as a pointwise MLP or a graph neural network, which allows the application of the VCNeF model to irregular grids.

\section{Conclusion}
In this work, we have designed an effective Neural PDE Solver, \emph{Vectorized Conditional Neural Field}, based on the conditional neural fields framework and demonstrated its generalization capabilities across multiple axes of desiderata: spatial, temporal, ICs, and PDE parameters. As a future work, we aim to experiment on turbulent simulations, improve the model design further with adaptive time-stepping, investigate sophisticated strategies for conditioning the neural fields~\cite{attention-vs-concat-neural-fields-conditioning-rebain:2023}, and test physics-informed losses. Additionally, we plan to investigate the effect of the temporal discretization the model was exposed to during training on the temporal zero-shot super-resolution capabilities of the model.

\section*{Impact Statement}
Accelerated and efficient Neural PDE Solvers help reduce costs in running cost-prohibitive simulations such as weather forecasting and cyclone predictions. Consequently, disaster preparedness, the design, and the manufacturing timeline can be accelerated, resulting in life and cost savings and decreased CO\textsubscript{2} emissions. As a negative side-effect, we cannot rule out the possibility of misuse by bad actors since fluid dynamic simulations are used to design military equipment.

\section*{Acknowledgements}
Funded by Deutsche Forschungsgemeinschaft (DFG, German Research Foundation) under Germany's Excellence Strategy - EXC 2075 – 390740016. We acknowledge the support of the Stuttgart Center for Simulation Science (SimTech). The authors thank the International Max Planck Research School for Intelligent Systems (IMPRS-IS) for supporting Marimuthu Kalimuthu, Daniel Musekamp, and Mathias Niepert. Additionally, we acknowledge the support of the German Federal Ministry of Education and Research (BMBF) as part of InnoPhase (funding code: 02NUK078). Furthermore, we acknowledge the support of the European Laboratory for Learning and Intelligent Systems (ELLIS) Unit Stuttgart. Lastly, we thank Makoto Takamoto for the insightful discussions.


\bibliography{references}

\begin{thebibliography}{42}
\providecommand{\natexlab}[1]{#1}
\providecommand{\url}[1]{\texttt{#1}}
\expandafter\ifx\csname urlstyle\endcsname\relax
  \providecommand{\doi}[1]{doi: #1}\else
  \providecommand{\doi}{doi: \begingroup \urlstyle{rm}\Url}\fi

\bibitem[Boussif et~al.(2022)Boussif, Bengio, Benabbou, and Assouline]{magnet-pde-boussif:2022}
Boussif, O., Bengio, Y., Benabbou, L., and Assouline, D.
\newblock Magnet: Mesh agnostic neural pde solver.
\newblock \emph{Advances in Neural Information Processing Systems}, 35:\penalty0 31972--31985, 2022.

\bibitem[Brandstetter et~al.(2022)Brandstetter, Worrall, and Welling]{mp-pde-brandstetter}
Brandstetter, J., Worrall, D.~E., and Welling, M.
\newblock Message passing neural {PDE} solvers.
\newblock \emph{CoRR}, abs/2202.03376, 2022.
\newblock URL \url{https://arxiv.org/abs/2202.03376}.

\bibitem[Cao(2021)]{galerkin-cao}
Cao, S.
\newblock Choose a transformer: Fourier or galerkin.
\newblock \emph{CoRR}, abs/2105.14995, 2021.
\newblock URL \url{https://arxiv.org/abs/2105.14995}.

\bibitem[Chen et~al.(2021)Chen, Fan, and Panda]{multi-scale-vit-chen}
Chen, C., Fan, Q., and Panda, R.
\newblock Crossvit: Cross-attention multi-scale vision transformer for image classification.
\newblock \emph{CoRR}, abs/2103.14899, 2021.
\newblock URL \url{https://arxiv.org/abs/2103.14899}.

\bibitem[Chen et~al.(2023{\natexlab{a}})Chen, Wu, Grinspun, Zheng, and Chen]{implicit-neural-spatial-repr-pde-chen:2023}
Chen, H., Wu, R., Grinspun, E., Zheng, C., and Chen, P.~Y.
\newblock Implicit neural spatial representations for time-dependent pdes.
\newblock In Krause, A., Brunskill, E., Cho, K., Engelhardt, B., Sabato, S., and Scarlett, J. (eds.), \emph{International Conference on Machine Learning, {ICML} 2023, 23-29 July 2023, Honolulu, Hawaii, {USA}}, volume 202 of \emph{Proceedings of Machine Learning Research}, pp.\  5162--5177. {PMLR}, 2023{\natexlab{a}}.
\newblock URL \url{https://proceedings.mlr.press/v202/chen23af.html}.

\bibitem[Chen et~al.(2023{\natexlab{b}})Chen, Xiang, Cho, Chang, Pershing, Maia, Chiaramonte, Carlberg, and Grinspun]{crom-pde-yichen-chen:2023}
Chen, P.~Y., Xiang, J., Cho, D.~H., Chang, Y., Pershing, G.~A., Maia, H.~T., Chiaramonte, M.~M., Carlberg, K.~T., and Grinspun, E.
\newblock {CROM:} continuous reduced-order modeling of pdes using implicit neural representations.
\newblock In \emph{The Eleventh International Conference on Learning Representations, {ICLR} 2023, Kigali, Rwanda, May 1-5, 2023}. OpenReview.net, 2023{\natexlab{b}}.
\newblock URL \url{https://openreview.net/pdf?id=FUORz1tG8Og}.

\bibitem[Chu et~al.(2022)Chu, Liu, Zheng, Franz, Seidel, Theobalt, and Zayer]{physics-informed-neural-fields-chu:2022}
Chu, M., Liu, L., Zheng, Q., Franz, E., Seidel, H., Theobalt, C., and Zayer, R.
\newblock Physics informed neural fields for smoke reconstruction with sparse data.
\newblock \emph{{ACM} Trans. Graph.}, 41\penalty0 (4):\penalty0 119:1--119:14, 2022.
\newblock \doi{10.1145/3528223.3530169}.
\newblock URL \url{https://doi.org/10.1145/3528223.3530169}.

\bibitem[Devlin et~al.(2018)Devlin, Chang, Lee, and Toutanova]{transfromers-bert-devlin}
Devlin, J., Chang, M., Lee, K., and Toutanova, K.
\newblock {BERT:} pre-training of deep bidirectional transformers for language understanding.
\newblock \emph{CoRR}, abs/1810.04805, 2018.
\newblock URL \url{http://arxiv.org/abs/1810.04805}.

\bibitem[Dosovitskiy et~al.(2020)Dosovitskiy, Beyer, Kolesnikov, Weissenborn, Zhai, Unterthiner, Dehghani, Minderer, Heigold, Gelly, Uszkoreit, and Houlsby]{transformers-images-dosovitskiy}
Dosovitskiy, A., Beyer, L., Kolesnikov, A., Weissenborn, D., Zhai, X., Unterthiner, T., Dehghani, M., Minderer, M., Heigold, G., Gelly, S., Uszkoreit, J., and Houlsby, N.
\newblock An image is worth 16x16 words: Transformers for image recognition at scale.
\newblock \emph{CoRR}, abs/2010.11929, 2020.
\newblock URL \url{https://arxiv.org/abs/2010.11929}.

\bibitem[Geneva \& Zabaras(2020)Geneva and Zabaras]{transformers-physical-systems-geneva:2020}
Geneva, N. and Zabaras, N.
\newblock Transformers for modeling physical systems.
\newblock \emph{CoRR}, abs/2010.03957, 2020.
\newblock URL \url{https://arxiv.org/abs/2010.03957}.

\bibitem[Gulati et~al.(2020)Gulati, Qin, Chiu, Parmar, Zhang, Yu, Han, Wang, Zhang, Wu, and Pang]{transformer-conformer-gulati}
Gulati, A., Qin, J., Chiu, C.-C., Parmar, N., Zhang, Y., Yu, J., Han, W., Wang, S., Zhang, Z., Wu, Y., and Pang, R.
\newblock Conformer: Convolution-augmented transformer for speech recognition, 2020.

\bibitem[Gupta \& Brandstetter(2023)Gupta and Brandstetter]{generalized-pde-modeling-gupta:2023}
Gupta, J.~K. and Brandstetter, J.
\newblock Towards multi-spatiotemporal-scale generalized {PDE} modeling.
\newblock \emph{Transactions on Machine Learning Research}, 2023.
\newblock ISSN 2835-8856.
\newblock URL \url{https://openreview.net/forum?id=dPSTDbGtBY}.

\bibitem[Hao et~al.(2023)Hao, Wang, Su, Ying, Dong, Liu, Cheng, Song, and Zhu]{gnot-operator-learn-hao:2023}
Hao, Z., Wang, Z., Su, H., Ying, C., Dong, Y., Liu, S., Cheng, Z., Song, J., and Zhu, J.
\newblock {GNOT:} {A} general neural operator transformer for operator learning.
\newblock In Krause, A., Brunskill, E., Cho, K., Engelhardt, B., Sabato, S., and Scarlett, J. (eds.), \emph{International Conference on Machine Learning, {ICML} 2023, 23-29 July 2023, Honolulu, Hawaii, {USA}}, volume 202 of \emph{Proceedings of Machine Learning Research}, pp.\  12556--12569. {PMLR}, 2023.
\newblock URL \url{https://proceedings.mlr.press/v202/hao23c.html}.

\bibitem[Katharopoulos et~al.(2020)Katharopoulos, Vyas, Pappas, and Fleuret]{linear-transformers-katharopoulos:2020}
Katharopoulos, A., Vyas, A., Pappas, N., and Fleuret, F.
\newblock Transformers are rnns: Fast autoregressive transformers with linear attention.
\newblock In \emph{Proceedings of the 37th International Conference on Machine Learning, {ICML} 2020, 13-18 July 2020, Virtual Event}, volume 119 of \emph{Proceedings of Machine Learning Research}, pp.\  5156--5165. {PMLR}, 2020.
\newblock URL \url{http://proceedings.mlr.press/v119/katharopoulos20a.html}.

\bibitem[Kovachki et~al.(2021)Kovachki, Li, Liu, Azizzadenesheli, Bhattacharya, Stuart, and Anandkumar]{neural-operators-kovachki}
Kovachki, N.~B., Li, Z., Liu, B., Azizzadenesheli, K., Bhattacharya, K., Stuart, A.~M., and Anandkumar, A.
\newblock Neural operator: Learning maps between function spaces.
\newblock \emph{CoRR}, abs/2108.08481, 2021.
\newblock URL \url{https://arxiv.org/abs/2108.08481}.

\bibitem[Krishnapriyan et~al.(2021)Krishnapriyan, Gholami, Zhe, Kirby, and Mahoney]{pinn-failure-modes-krishnapriyan:2021}
Krishnapriyan, A.~S., Gholami, A., Zhe, S., Kirby, R.~M., and Mahoney, M.~W.
\newblock Characterizing possible failure modes in physics-informed neural networks.
\newblock In Ranzato, M., Beygelzimer, A., Dauphin, Y.~N., Liang, P., and Vaughan, J.~W. (eds.), \emph{Advances in Neural Information Processing Systems 34: Annual Conference on Neural Information Processing Systems 2021, NeurIPS 2021, December 6-14, 2021, virtual}, pp.\  26548--26560, 2021.
\newblock URL \url{https://arxiv.org/abs/2109.01050}.

\bibitem[Li et~al.(2021)Li, Si, Li, Hsieh, and Bengio]{fourier-features-li}
Li, Y., Si, S., Li, G., Hsieh, C.-J., and Bengio, S.
\newblock Learnable fourier features for multi-dimensional spatial positional encoding, 2021.

\bibitem[Li et~al.(2020{\natexlab{a}})Li, Kovachki, Azizzadenesheli, Liu, Bhattacharya, Stuart, and Anandkumar]{fno-li}
Li, Z., Kovachki, N.~B., Azizzadenesheli, K., Liu, B., Bhattacharya, K., Stuart, A.~M., and Anandkumar, A.
\newblock Fourier neural operator for parametric partial differential equations.
\newblock \emph{CoRR}, abs/2010.08895, 2020{\natexlab{a}}.
\newblock URL \url{https://arxiv.org/abs/2010.08895}.

\bibitem[Li et~al.(2020{\natexlab{b}})Li, Kovachki, Azizzadenesheli, Liu, Bhattacharya, Stuart, and Anandkumar]{graph-neural-operator-li}
Li, Z., Kovachki, N.~B., Azizzadenesheli, K., Liu, B., Bhattacharya, K., Stuart, A.~M., and Anandkumar, A.
\newblock Neural operator: Graph kernel network for partial differential equations.
\newblock \emph{CoRR}, abs/2003.03485, 2020{\natexlab{b}}.
\newblock URL \url{https://arxiv.org/abs/2003.03485}.

\bibitem[Li et~al.(2023{\natexlab{a}})Li, Meidani, and Farimani]{oformer-pde-li:2023}
Li, Z., Meidani, K., and Farimani, A.~B.
\newblock Transformer for partial differential equations' operator learning.
\newblock \emph{Trans. Mach. Learn. Res.}, 2023, 2023{\natexlab{a}}.
\newblock URL \url{https://openreview.net/forum?id=EPPqt3uERT}.

\bibitem[Li et~al.(2023{\natexlab{b}})Li, Shu, and Farimani]{scalable-transformer-pde-li:2023}
Li, Z., Shu, D., and Farimani, A.~B.
\newblock Scalable transformer for {PDE} surrogate modeling.
\newblock \emph{CoRR}, abs/2305.17560, 2023{\natexlab{b}}.
\newblock \doi{10.48550/arXiv.2305.17560}.
\newblock URL \url{https://doi.org/10.48550/arXiv.2305.17560}.

\bibitem[Li et~al.(2023{\natexlab{c}})Li, Zheng, Kovachki, Jin, Chen, Liu, Azizzadenesheli, and Anandkumar]{physics-neural-operator-li}
Li, Z., Zheng, H., Kovachki, N., Jin, D., Chen, H., Liu, B., Azizzadenesheli, K., and Anandkumar, A.
\newblock Physics-informed neural operator for learning partial differential equations, 2023{\natexlab{c}}.

\bibitem[Lippe et~al.(2023)Lippe, Veeling, Perdikaris, Turner, and Brandstetter]{pde-refiner-long-term-rollouts-neural-pde-lippe:2023}
Lippe, P., Veeling, B.~S., Perdikaris, P., Turner, R.~E., and Brandstetter, J.
\newblock Pde-refiner: Achieving accurate long rollouts with neural {PDE} solvers.
\newblock \emph{CoRR}, abs/2308.05732, 2023.
\newblock \doi{10.48550/ARXIV.2308.05732}.
\newblock URL \url{https://doi.org/10.48550/arXiv.2308.05732}.

\bibitem[Lu et~al.(2019)Lu, Jin, and Karniadakis]{deeponet-lu:2019}
Lu, L., Jin, P., and Karniadakis, G.~E.
\newblock Deeponet: Learning nonlinear operators for identifying differential equations based on the universal approximation theorem of operators.
\newblock \emph{CoRR}, abs/1910.03193, 2019.
\newblock URL \url{http://arxiv.org/abs/1910.03193}.

\bibitem[Maclaurin et~al.(2015)Maclaurin, Duvenaud, and Adams]{hips-autograd-maclaurin:2015}
Maclaurin, D., Duvenaud, D., and Adams, R.~P.
\newblock Autograd: Effortless gradients in numpy.
\newblock In \emph{ICML 2015 AutoML workshop}, volume 238, 2015.

\bibitem[McCabe et~al.(2023)McCabe, Blancard, Parker, Ohana, Cranmer, Bietti, Eickenberg, Golkar, Krawezik, Lanusse, Pettee, Tesileanu, Cho, and Ho]{multi-physics-pretrain-avit-mccabe:2023}
McCabe, M., Blancard, B.~R., Parker, L.~H., Ohana, R., Cranmer, M.~D., Bietti, A., Eickenberg, M., Golkar, S., Krawezik, G., Lanusse, F., Pettee, M., Tesileanu, T., Cho, K., and Ho, S.
\newblock Multiple physics pretraining for physical surrogate models.
\newblock \emph{CoRR}, abs/2310.02994, 2023.
\newblock \doi{10.48550/ARXIV.2310.02994}.
\newblock URL \url{https://doi.org/10.48550/arXiv.2310.02994}.

\bibitem[Mescheder et~al.(2019)Mescheder, Oechsle, Niemeyer, Nowozin, and Geiger]{occupancy-networks-inr-mescheder:2019}
Mescheder, L.~M., Oechsle, M., Niemeyer, M., Nowozin, S., and Geiger, A.
\newblock Occupancy networks: Learning 3d reconstruction in function space.
\newblock In \emph{{IEEE} Conference on Computer Vision and Pattern Recognition, {CVPR} 2019, Long Beach, CA, USA, June 16-20, 2019}, pp.\  4460--4470. Computer Vision Foundation / {IEEE}, 2019.
\newblock \doi{10.1109/CVPR.2019.00459}.
\newblock URL \url{http://openaccess.thecvf.com/content\_CVPR\_2019/html/Mescheder\_Occupancy\_Networks\_Learning\_3D\_Reconstruction\_in\_Function\_Space\_CVPR\_2019\_paper.html}.

\bibitem[Ovadia et~al.(2023)Ovadia, Turkel, Kahana, and Karniadakis]{ditto-diffusion-temporal-transformer-ovadia:2023}
Ovadia, O., Turkel, E., Kahana, A., and Karniadakis, G.~E.
\newblock Ditto: Diffusion-inspired temporal transformer operator.
\newblock \emph{CoRR}, abs/2307.09072, 2023.
\newblock \doi{10.48550/ARXIV.2307.09072}.
\newblock URL \url{https://doi.org/10.48550/arXiv.2307.09072}.

\bibitem[Paszke et~al.(2017)Paszke, Gross, Chintala, Chanan, Yang, DeVito, Lin, Desmaison, Antiga, and Lerer]{autodiff-pytorch-paszke:2017}
Paszke, A., Gross, S., Chintala, S., Chanan, G., Yang, E., DeVito, Z., Lin, Z., Desmaison, A., Antiga, L., and Lerer, A.
\newblock Automatic differentiation in pytorch.
\newblock In \emph{NIPS 2017 Workshop on Autodiff}, 2017.
\newblock URL \url{https://openreview.net/forum?id=BJJsrmfCZ}.

\bibitem[Perez et~al.(2018)Perez, Strub, de~Vries, Dumoulin, and Courville]{film-conditioning-perez:2018}
Perez, E., Strub, F., de~Vries, H., Dumoulin, V., and Courville, A.~C.
\newblock Film: Visual reasoning with a general conditioning layer.
\newblock In McIlraith, S.~A. and Weinberger, K.~Q. (eds.), \emph{Proceedings of the Thirty-Second {AAAI} Conference on Artificial Intelligence, (AAAI-18), the 30th innovative Applications of Artificial Intelligence (IAAI-18), and the 8th {AAAI} Symposium on Educational Advances in Artificial Intelligence (EAAI-18), New Orleans, Louisiana, USA, February 2-7, 2018}, pp.\  3942--3951. {AAAI} Press, 2018.
\newblock \doi{10.1609/AAAI.V32I1.11671}.
\newblock URL \url{https://doi.org/10.1609/aaai.v32i1.11671}.

\bibitem[Rahman et~al.(2023)Rahman, Ross, and Azizzadenesheli]{unet-neural-op-uno-rahman:2023}
Rahman, M.~A., Ross, Z.~E., and Azizzadenesheli, K.
\newblock U-{NO}: U-shaped neural operators.
\newblock \emph{Transactions on Machine Learning Research}, 2023.
\newblock ISSN 2835-8856.
\newblock URL \url{https://openreview.net/forum?id=j3oQF9coJd}.

\bibitem[Raissi et~al.(2017)Raissi, Perdikaris, and Karniadakis]{pinn-raissi}
Raissi, M., Perdikaris, P., and Karniadakis, G.~E.
\newblock Physics informed deep learning (part {I):} data-driven solutions of nonlinear partial differential equations.
\newblock \emph{CoRR}, abs/1711.10561, 2017.
\newblock URL \url{http://arxiv.org/abs/1711.10561}.

\bibitem[Rebain et~al.(2023)Rebain, Matthews, Yi, Sharma, Lagun, and Tagliasacchi]{attention-vs-concat-neural-fields-conditioning-rebain:2023}
Rebain, D., Matthews, M.~J., Yi, K.~M., Sharma, G., Lagun, D., and Tagliasacchi, A.
\newblock Attention beats concatenation for conditioning neural fields.
\newblock \emph{Transactions on Machine Learning Research}, 2023.
\newblock ISSN 2835-8856.
\newblock URL \url{https://openreview.net/forum?id=GzqdMrFQsE}.

\bibitem[Ruoss et~al.(2023)Ruoss, Delétang, Genewein, Grau-Moya, Csordás, Bennani, Legg, and Veness]{rope-ruoss}
Ruoss, A., Delétang, G., Genewein, T., Grau-Moya, J., Csordás, R., Bennani, M., Legg, S., and Veness, J.
\newblock Randomized positional encodings boost length generalization of transformers, 2023.

\bibitem[Sanchez{-}Gonzalez et~al.(2020)Sanchez{-}Gonzalez, Godwin, Pfaff, Ying, Leskovec, and Battaglia]{learn2simulate-physics-sanchez-gonzalez:2020}
Sanchez{-}Gonzalez, A., Godwin, J., Pfaff, T., Ying, R., Leskovec, J., and Battaglia, P.~W.
\newblock Learning to simulate complex physics with graph networks.
\newblock In \emph{Proceedings of the 37th International Conference on Machine Learning, {ICML} 2020, 13-18 July 2020, Virtual Event}, volume 119 of \emph{Proceedings of Machine Learning Research}, pp.\  8459--8468. {PMLR}, 2020.
\newblock URL \url{http://proceedings.mlr.press/v119/sanchez-gonzalez20a.html}.

\bibitem[Serrano et~al.(2023)Serrano, Boudec, Koupa{\"{\i}}, Wang, Yin, Vittaut, and Gallinari]{operator-learning-neural-fields-general-geometries-serrano:2023}
Serrano, L., Boudec, L.~L., Koupa{\"{\i}}, A.~K., Wang, T.~X., Yin, Y., Vittaut, J., and Gallinari, P.
\newblock Operator learning with neural fields: Tackling pdes on general geometries.
\newblock \emph{CoRR}, abs/2306.07266, 2023.
\newblock \doi{10.48550/ARXIV.2306.07266}.
\newblock URL \url{https://doi.org/10.48550/arXiv.2306.07266}.

\bibitem[Sitzmann et~al.(2020)Sitzmann, Martel, Bergman, Lindell, and Wetzstein]{siren-inr-sitzmann:2020}
Sitzmann, V., Martel, J. N.~P., Bergman, A.~W., Lindell, D.~B., and Wetzstein, G.
\newblock Implicit neural representations with periodic activation functions.
\newblock In Larochelle, H., Ranzato, M., Hadsell, R., Balcan, M., and Lin, H. (eds.), \emph{Advances in Neural Information Processing Systems 33: Annual Conference on Neural Information Processing Systems 2020, NeurIPS 2020, December 6-12, 2020, virtual}, 2020.
\newblock URL \url{https://proceedings.neurips.cc/paper/2020/hash/53c04118df112c13a8c34b38343b9c10-Abstract.html}.

\bibitem[Takamoto et~al.(2022)Takamoto, Praditia, Leiteritz, MacKinlay, Alesiani, Pfl{\"u}ger, and Niepert]{pdebench-takamoto:2022}
Takamoto, M., Praditia, T., Leiteritz, R., MacKinlay, D., Alesiani, F., Pfl{\"u}ger, D., and Niepert, M.
\newblock Pdebench: An extensive benchmark for scientific machine learning.
\newblock \emph{NeurIPS}, 2022.

\bibitem[Takamoto et~al.(2023)Takamoto, Alesiani, and Niepert]{cape-takamoto:2023}
Takamoto, M., Alesiani, F., and Niepert, M.
\newblock Learning neural {PDE} solvers with parameter-guided channel attention.
\newblock \emph{ICML}, abs/2304.14118, 2023.
\newblock \doi{10.48550/arXiv.2304.14118}.
\newblock URL \url{https://doi.org/10.48550/arXiv.2304.14118}.

\bibitem[Vaswani et~al.(2017)Vaswani, Shazeer, Parmar, Uszkoreit, Jones, Gomez, Kaiser, and Polosukhin]{transformers-vaswani}
Vaswani, A., Shazeer, N., Parmar, N., Uszkoreit, J., Jones, L., Gomez, A.~N., Kaiser, L., and Polosukhin, I.
\newblock Attention is all you need.
\newblock \emph{CoRR}, abs/1706.03762, 2017.
\newblock URL \url{http://arxiv.org/abs/1706.03762}.

\bibitem[Xie et~al.(2021)Xie, Takikawa, Saito, Litany, Yan, Khan, Tombari, Tompkin, Sitzmann, and Sridhar]{neural-fields-xie}
Xie, Y., Takikawa, T., Saito, S., Litany, O., Yan, S., Khan, N., Tombari, F., Tompkin, J., Sitzmann, V., and Sridhar, S.
\newblock Neural fields in visual computing and beyond.
\newblock \emph{CoRR}, abs/2111.11426, 2021.
\newblock URL \url{https://arxiv.org/abs/2111.11426}.

\bibitem[Yin et~al.(2023)Yin, Kirchmeyer, Franceschi, Rakotomamonjy, and Gallinari]{dynamics-aware-implicit-neural-repr-dino-yin:2023}
Yin, Y., Kirchmeyer, M., Franceschi, J., Rakotomamonjy, A., and Gallinari, P.
\newblock Continuous {PDE} dynamics forecasting with implicit neural representations.
\newblock In \emph{The Eleventh International Conference on Learning Representations, {ICLR} 2023, Kigali, Rwanda, May 1-5, 2023}. OpenReview.net, 2023.
\newblock URL \url{https://openreview.net/pdf?id=B73niNjbPs}.

\end{thebibliography}
\bibliographystyle{icml2024}

\newpage
\appendix
\onecolumn

\begin{center}
\startcontents[sections]\vbox{\vspace{3mm}\sc \LARGE Vectorized Conditional Neural Fields: A Framework for Solving Time-dependent Partial Differential Equations\\\sc\LARGE Appendix \\
\Large \textbf{Code}: {\href{https://github.com/jhagnberger/vcnef/}{\textbf{https://github.com/jhagnberger/vcnef/}}}
}\vspace{4mm}\hrule height .5pt
\end{center}
\printcontents[sections]{l}{0}{\setcounter{tocdepth}{2}}
\newpage

\section{Comparison of Neural Architectures for PDE Solving}
\label{app:comparison-of-architectures}

\cref{tab:model-features} shows the most important properties of ML models for solving PDEs and compares three families of models. The VCNeF combines both worlds of neural fields (e.g., PINNs) and neural operators. PINNs do not leverage the spatial dependencies among the queried coordinates since they produce the output for each queried coordinate independently. Meanwhile, neural operators map to a set of solution points and leverage the dependencies between the regressed points. However, neural operator implementations usually have limited support for time continuity, while PINNs are time-continuous. VCNeF, therefore, combines the advantages of both worlds by leveraging spatial dependencies, generating a set of solution points, and being continuous in time.

\begin{table}[!htb]
    \centering
    \scalebox{0.85}{
        \begin{tabular}{ ccccccc }
            \toprule
            \multirow{2}{*}{Model family} & \multirow{2}{*}{Model} & Initial value & PDE parameter & \multicolumn{2}{c}{ZSSR} & Models spatial dependencies \\
            & & generalization & generalization & Spatial & Temporal & with self-attention \\
            \midrule
            \multirow{5}{*}{Neural Operator} & FNO$^1$ & \cmark & \xmark & \cmark & \xmark & \xmark \\
            & OFormer & \cmark & \xmark  & \cmark & \xmark & \cmark \\
            \cmidrule{2-7}
            & cFNO$^1$ & \cmark & \cmark &  \cmark & \xmark & \xmark \\
            & cOFormer & \cmark & \cmark & \cmark & \xmark & \cmark \\
            \midrule
            Neural Field & PINN & \xmark & \xmark & \cmark & \cmark & \xmark \\
            \midrule
            \multirow{2}{*}{Conditional Neural Field}  & CORAL & \cmark & \xmark  & \cmark & \cmark & \xmark \\
            & VCNeF (ours) & \cmark & \cmark & \cmark & \cmark & \cmark \\
            \bottomrule
        \end{tabular}
    }
    \caption{Overview of distinct properties of benchmark baselines and our proposed VCNeF model. $^1$Refers to FNO that applies the Fourier transform only to the spatial domain and is trained in an autoregressive fashion.}
    \label{tab:model-features}
\end{table}

\section{Additional VCNeF Model Details}

\subsection{Vectorized Conditional Neural Field as a Neural Operator}
\label{app:vcnef-as-no}

Neural operators are theoretically time-continuous. However, current neural operator implementations have limited support for being time continuous. Our proposed architecture for 1D PDEs can be considered as a neural operator implementation that is conditioned on time to be temporally continuous. As a time-continuous neural operator implementation, VCNeF learns a mapping between the initial condition (i.e., input function $u(0, \cdot)$) and the solution at time $t$ (i.e., output function $u(t, \cdot)$). This can mathematically be expressed as

\begin{equation}
    f_\theta(u(0, \boldsymbol{x}))(t) \approx u(t, \boldsymbol{x})
    \label{eq:vcnef-as-neural-operator}
\end{equation}

where $f_\theta$ denotes the neural network or neural operator. The VCNeF model can be decomposed into the following layers that are identical to the Fourier Neural Operator of \citet{fno-li}.

\paragraph{Lifting.}
Encoding the input functions with a shared pointwise linear layer represents a lifting of the input function $u^{(0)}(\boldsymbol{x}) := u(0, \boldsymbol{x})$ that lifts the function to a higher dimensional space.

\begin{equation}
    u^{(1)} = u^{(0)}\boldsymbol{W} + \boldsymbol{b}
    \label{eq:vcnef-neural-operator-lifting}
\end{equation}

which can be equivalently represented as a function

\begin{equation}
    u^{(1)}(\boldsymbol{x}) = (u^{(0)}\boldsymbol{W} + \boldsymbol{b})(\boldsymbol{x}) \qquad \forall \boldsymbol{x} \in \mathbb{X} 
    \label{eq:vcnef-neural-operator-lifting-func}
\end{equation}

\paragraph{Iterative Updates.} 
Following \citet{galerkin-cao} and \citet{neural-operators-kovachki} and interpreting the columns $j$ of the matrices $\boldsymbol{Q}, \boldsymbol{K}, \boldsymbol{V}$ as learnable basis functions $q_j, k_j, v_j$ yields that the encoding of the initial condition with self-attention of the Linear Transformer blocks can be seen as a kernel integral transformation as follows

\begin{equation}
    \begin{gathered}
    Q = u^{(n)}W_Q + b_Q \qquad K = u^{(n)}W_K + b_K \qquad V = u^{(n)}W_V + b_V \\
    u^{(n+1)} = \mathtt{Linear\_Attn}(Q, K, V) = \frac{\Phi(Q)\Phi(K)^\top V}{\Phi(Q)^\top \Phi(K)} = \frac{1}{\Phi(Q)^\top \Phi(K)} \Phi(Q)\Phi(K)^\top V \\
    u^{(n+1)}_{i,j} = \frac{1}{\sum_{l=1}^s\Phi(Q_i)^\top \Phi(K_l)} \sum_{l = 1}^s \left( \Phi(Q_i)^\top \Phi(K_l) \right) V_{l, j} \approx \int_\mathbb{X} \frac{\kappa(\boldsymbol{x_i}, \xi)}{\int_\mathbb{X} \kappa(\boldsymbol{x_i}, \psi) d\psi} v_j(\xi) d\xi
    \end{gathered}
    \label{eq:vcnef-kernel-integral}
\end{equation}

where the learnable kernel $\kappa(\boldsymbol{x_i}, \xi)$ is approximated by $\Phi(Q_i)^\top \Phi(K_l)$. It can also be expressed as a function

\begin{equation}
    u^{(n+1)}(\boldsymbol{x}) = \int_\mathbb{X} \frac{\kappa(\boldsymbol{x}, \xi)}{{\int_\mathbb{X} \kappa(\boldsymbol{x}, \psi) d\psi}} (u^{(n)}W_V+b_V)(\xi) d\xi  \qquad \forall \boldsymbol{x} \in \mathbb{X}
    \label{eq:vcnef-kernel-integral-func}
\end{equation}

\paragraph{Iterative Time Injection.}
The modulation blocks can be considered as a time injection mechanism that performs the kernel integral transformation from \cref{eq:vcnef-kernel-integral} and multiplies the latent representation $u^{(m+1)}$ with a spatial and temporal dependent function $g$.

\begin{equation}
    \begin{gathered}
    g := g(t, \boldsymbol{x}) = \mathtt{MLP}(t \mathbin\Vert \boldsymbol{x}) \\
    u^{(m+1)} = \mathtt{Modulation\_Block}(g, u^{(m)}) = \sigma \left( \mathtt{Linear\_Attn}(Q, K, V) \right) \circ g \\
    \text{ with } Q = u^{(m)}W_Q + b_Q \qquad K = u^{(m)}W_K + b_K \qquad V = u^{(m)}W_V + b_V \\
    u^{(m+1)}_{i, j} = \sigma \left( \int_\mathbb{X} \frac{\kappa(\boldsymbol{x_i}, \xi)}{\int_\mathbb{X} \kappa(\boldsymbol{x_i}, \psi) d\psi} v_j(\xi) d\xi \right) g_j(t, \boldsymbol{x_i})
    \end{gathered}
    \label{eq:vcnef-time-injection}
\end{equation}

which can also be represented as a function

\begin{equation}
    \begin{gathered}
    g := g(t, \boldsymbol{x}) = \mathtt{MLP}(t \mathbin\Vert \boldsymbol{x}) \\
    u^{(m+1)}(t, \boldsymbol{x}) = \sigma \left( \int_\mathbb{X} \frac{\kappa(\boldsymbol{x}, \xi)}{{\int_\mathbb{X} \kappa(\boldsymbol{x}, \psi) d\psi}} (u^{(m)}W_V+b_V)(\xi) d\xi  \right) g(t, \boldsymbol{x}) \qquad \forall \boldsymbol{x} \in \mathbb{X}, t \in (0, T] \\
    \end{gathered}
    \label{eq:vcnef-time-injection-func}
\end{equation}

\paragraph{Projection.}
The final hidden representation, which represents the solution $u(t, \boldsymbol{x})$, is projected back to the physical space with a pointwise MLP.

\begin{equation}
    u = u^{(m+n)}\boldsymbol{W} + \boldsymbol{b}
    \label{eq:vcnef-neural-operator-proj}
\end{equation}

which can be equivalently represented as a function

\begin{equation}
    u(\boldsymbol{x}) = (u^{(m+n)}\boldsymbol{W} + \boldsymbol{b})(\boldsymbol{x}) \qquad \forall \boldsymbol{x} \in \mathbb{X} 
    \label{eq:vcnef-neural-operator-proj-func}
\end{equation}

\subsection{Neural Architecture}

\begin{figure}[!htb]
    \centering
    \begin{minipage}{.325\textwidth}
        \centering
        \includegraphics[width=1\textwidth]{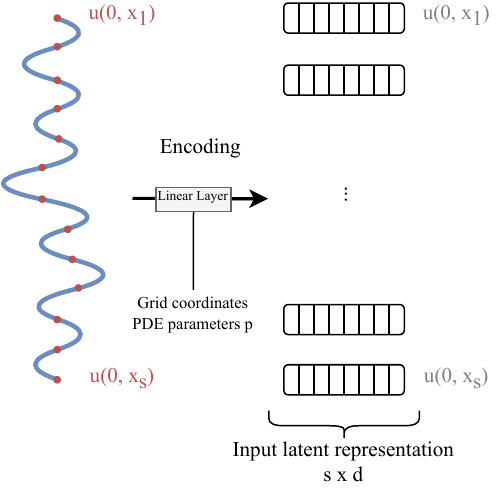}
    \end{minipage}%
    \hfill
    \begin{minipage}{0.65\textwidth}
        \centering
        \includegraphics[width=1\textwidth]{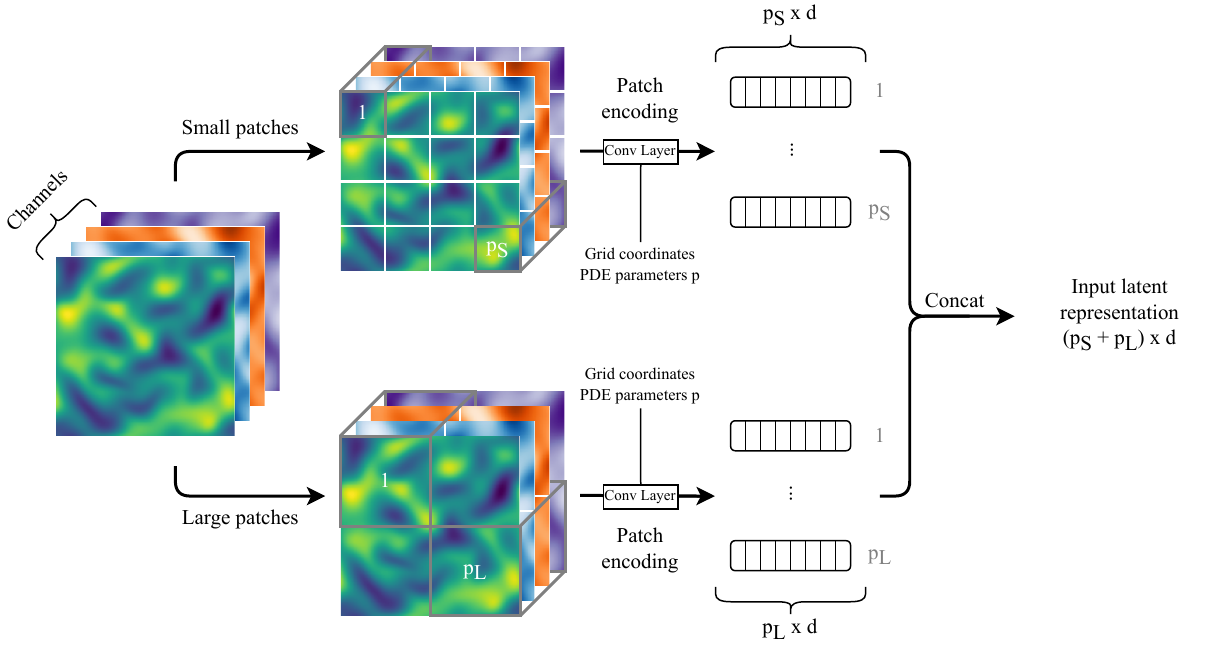}
    \end{minipage}
    \begin{minipage}[t]{.3\textwidth}
        \centering
        \subcaption{Latent representation for 1D PDEs}
        \label{subfig:encoding-latent-1d}
    \end{minipage}%
    \hfill
    \begin{minipage}[t]{0.63\textwidth}
        \centering
        \subcaption{Latent representation for 2D PDEs (multi-scale patches)}
        \label{subfig:encoding-latent-2d}
    \end{minipage}
    \caption{Encoding mechanisms for the initial condition of 1D and 2D PDEs.}
    \label{fig:encoding-latent-representations}
\end{figure}

\paragraph{Encoding of the Initial Condition.}
Depending on the dimensionality of the PDE, different mechanisms are utilized to generate latent representations of the IC. For 1D PDEs (Figure \ref{subfig:encoding-latent-1d}), each solution point in space is projected to a latent representation by applying a linear layer that is shared across the spatial points. For 2D PDEs (Figure \ref{subfig:encoding-latent-2d}), the initial condition is divided into patches of different sizes, and each patch is projected to a latent representation by a 2D convolutional layer. The IC of 3D PDEs is also divided into non-overlapping patches to reduce the computational costs. The latent representations of the small and large patches are concatenated together. $p_S$ denotes the number of small patches and depends on the spatial resolution of the input as well as the size of the patch. The same is true for $p_L$, which denotes the large patches. In addition to the initial condition, the encoding mechanism takes the grid as positional information and the PDE parameters $\boldsymbol{p}$ as input. The 1D case could also be seen as a special case of the patches with a 1D input and only one patch size of $1 \times 1$.

\begin{figure}[!htb]
    \begin{center}
        \includegraphics[width=0.7\textwidth]{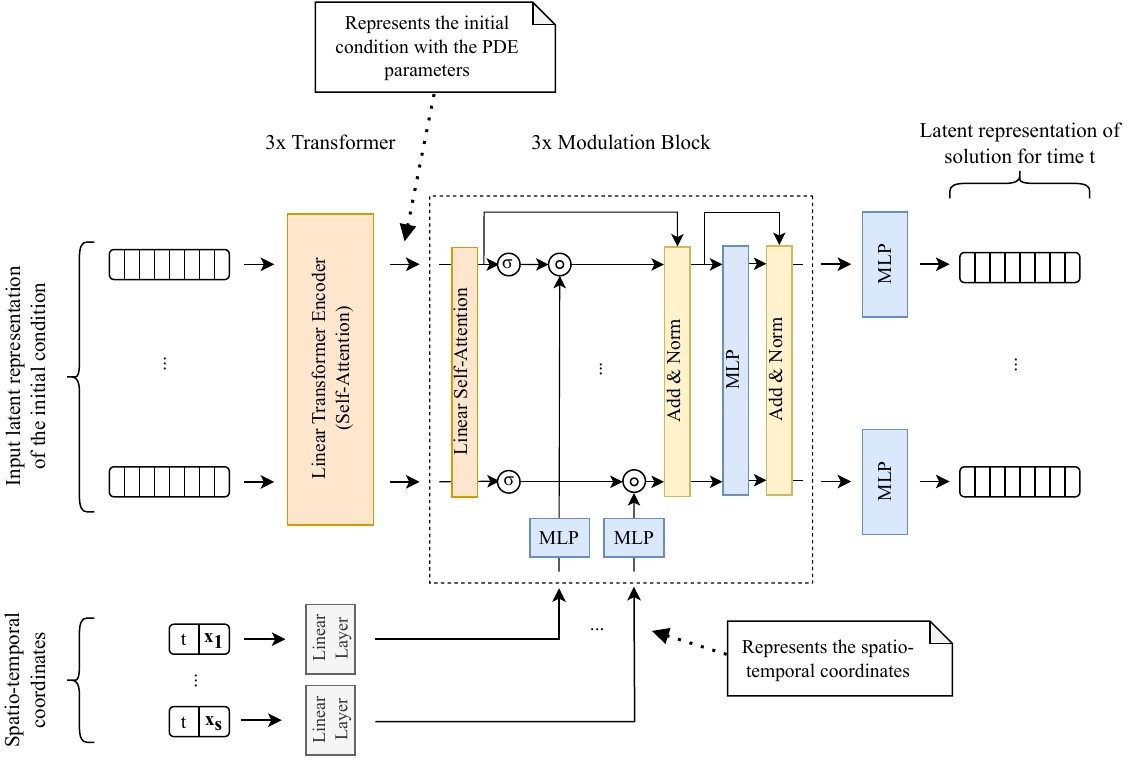}
        \caption{Architecture of VCNeF for solving time-dependent PDEs. The input latent representation is generated via the mechanisms in Figure \ref{fig:encoding-latent-representations}. The modulation block (dashed rectangle) contains a non-linearity $\sigma$, linear self-attention, two shared MLPs, and a pointwise multiplication $\circ$ (scaling of FiLM). Additionally, it contains residual connections and layer normalization (Add \& Norm). The solution's latent representation is mapped to physical space with the mechanism in Figure \ref{fig:vcnef-decoding}.}
        \label{fig:vcnef-architecture-detailed}
    \end{center}
\end{figure}

\paragraph{Transformer-based Vectorized Conditional Neural Field.}
\cref{fig:vcnef-architecture-detailed} shows the detailed architecture of the proposed VCNeF with the modulation blocks that modulate the latent representation of the input coordinates based on the IC. Linear self-attention allows an information flow between different spatio-temporal coordinates to capture spatial dependencies.

\begin{figure}[!htb]
    \begin{center}
        \includegraphics[width=0.75\textwidth]{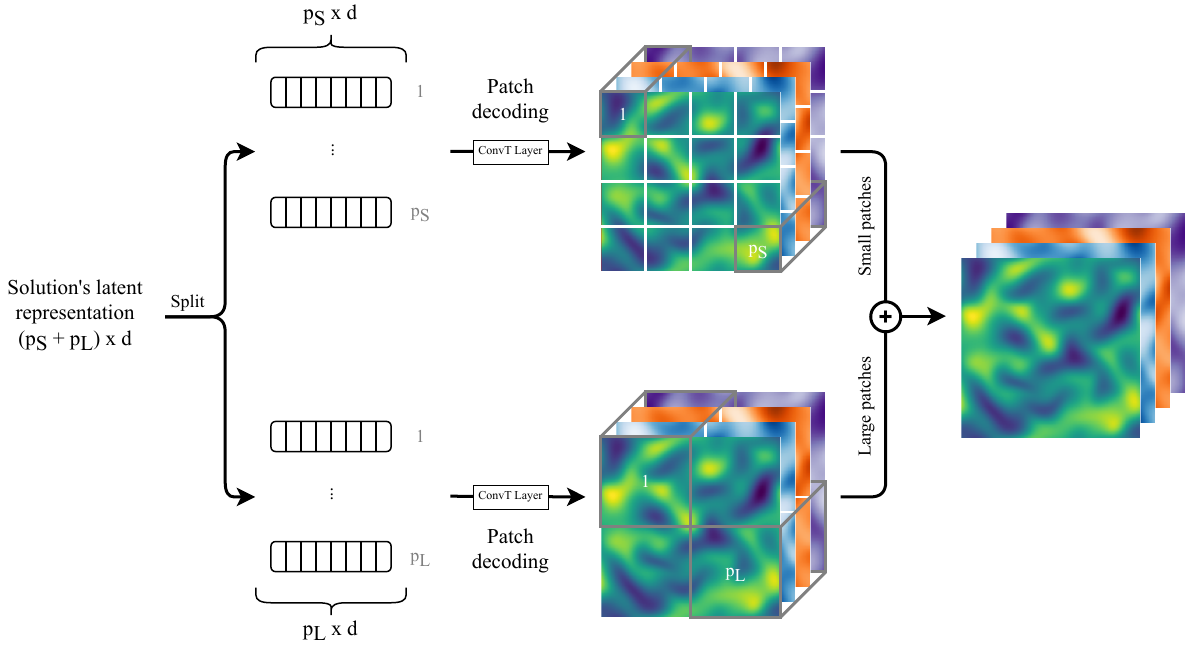}
        \caption{Solution decoding of the VCNeF for solving 2D PDEs. The solution's latent representation is split into the latent representation for the small and large patches. Thereafter, the latent representations are projected to patches with shared 2D convolution transposed layers. The final output is the weighted sum of the small and large patches.}
        \label{fig:vcnef-decoding}
    \end{center}
\end{figure}

\paragraph{Decoding of Solution's Latent Representation.}
Similar to the encoding, the decoding of the solution's latent representation depends on the dimensionality of the PDE. For 1D PDEs, a shared MLP is applied to decode the latent representation into the physical representation. \cref{fig:vcnef-decoding} shows the mechanism to map the solution's latent representation back to the physical space for a 2D PDE. The mechanism for 3D PDEs is similar to the mechanism for 2D PDEs, except that it operates on 3D patches instead of 2D patches.

\subsection{Linear Attention}
\citet{linear-transformers-katharopoulos:2020} reformulate the attention calculation to
\begin{equation}
    \mathtt{Attn}(\boldsymbol{Q}, \boldsymbol{K}, \boldsymbol{V}) = \mathtt{softmax}\big(\frac{\boldsymbol{QK}^{\top}}{\sqrt{d}}\big)\boldsymbol{V} = \boldsymbol{V'}
    \Leftrightarrow
    \boldsymbol{V'}_i = \frac{\sum_{j=1}^N\mathtt{sim}(\boldsymbol{Q}_i, \boldsymbol{K}_j) \boldsymbol{V}_j}{\sum_{j=1}^N\mathtt{sim}(\boldsymbol{Q}_i, \boldsymbol{K}_j)}
    \label{eq:attention-reformulated}
\end{equation}

where $\boldsymbol{V'}_i$ denotes the $i$-th row of matrix $\boldsymbol{V}$ and $\mathtt{sim}(\boldsymbol{q}, \boldsymbol{k}) = \text{exp}(\frac{\boldsymbol{q}^{\top}\boldsymbol{k}}{\sqrt{d}})$. $\mathtt{sim}(\cdot, \cdot)$ is a similarity function that measures the similarity of two vectors. Each function that takes two vectors as an input and outputs a non-negative real value has an interpretation as a similarity function. All kernel functions $\kappa(\boldsymbol{x}, \boldsymbol{y}): (\mathbb{R}^d \times \mathbb{R}^d) \rightarrow \mathbb{R}_{+}$ with non-negative output values also satisfies this property. Since kernel functions can be rewritten as the dot product of two output vectors of a feature function $\Phi(\cdot)$ that maps the input to some high-dimensional space (Mercer’s theorem), the kernel function can be written as $\kappa(\boldsymbol{x}, \boldsymbol{y}) = \Phi(\boldsymbol{x})^{\top} \Phi(\boldsymbol{y})$. This leads to a new interpretation of the attention equation:

\begin{subequations}
    \begin{equation}
        \boldsymbol{V'}_i = \frac{\sum_{j=1}^N \eqnmarkbox[indianred]{sim11}{ \mathtt{sim}(\boldsymbol{Q}_i, \boldsymbol{K}_j)} \boldsymbol{V}_j}{\sum_{j=1}^N \eqnmarkbox[indianred]{sim21}{ \mathtt{sim}(\boldsymbol{Q}_i, \boldsymbol{K}_j) }} 
        = \frac{\sum_{j=1}^N \eqnmarkbox[indianred]{sim12}{ \Phi(\boldsymbol{Q}_i)^\top \Phi(\boldsymbol{K}_j)} \boldsymbol{V}_j}{\sum_{j=1}^N \eqnmarkbox[indianred]{sim22}{ \Phi(\boldsymbol{Q}_i)^\top \Phi(\boldsymbol{K}_j) }}
        = \frac{ \eqnmarkbox[indianred]{pi21}{\Phi(\boldsymbol{Q}_i)^{\top}} \sum_{j=1}^N \eqnmarkbox[indianred]{sim12}{\Phi(\boldsymbol{K}_j)} \boldsymbol{V}_j^{\top}}{ \eqnmarkbox[indianred]{sim21}{\Phi(\boldsymbol{Q}_i)^{\top}} \sum_{j=1}^N \eqnmarkbox[indianred]{sim22}{\Phi(\boldsymbol{K}_j)}}
        \label{eq:attention-reformulated-feature-function}
    \end{equation}
\end{subequations}

Since $\Phi(\boldsymbol{Q}_i)^\top$ is independent of the index $i$ of the sum, the associative property of matrix multiplication can be applied, and $\Phi(\boldsymbol{Q}_i)^\top$ can be pulled in front of the sum. The value of $\sum_{j=1}^N\Phi(\boldsymbol{K}_j) \boldsymbol{V}_j^\top$ and $\sum_{j=1}^N\Phi(\boldsymbol{K}_j)$ needs only to be calculated once because they can be reused (time complexity of $\mathcal{O}(N)$) and $\boldsymbol{V}_i$ needs to be calculated for all $N$ tokens (time complexity of $\mathcal{O}(N)$). This results in linear time and space complexity. \cref{eq:attention-reformulated-feature-function} can also be written in matrix notation:

\begin{equation}
    \mathtt{Linear\_Attn}(\boldsymbol{Q}, \boldsymbol{K}, \boldsymbol{V}) = \frac{\left(\Phi(\boldsymbol{Q})\Phi(\boldsymbol{K})^{\top} \right)\boldsymbol{V}}{\Phi(\boldsymbol{Q})^\top \Phi(\boldsymbol{K})} = \frac{\Phi(\boldsymbol{Q})\left(\Phi(\boldsymbol{K})^\top \boldsymbol{V}\right)}{\Phi(\boldsymbol{Q})^\top \Phi(\boldsymbol{K})}
    \label{eq:attention-matrix}
\end{equation}

where the feature function $\Phi$ is applied row-wise to the matrix.

\paragraph{Feature Function.}
We use the feature function
\begin{equation}
    \begin{aligned}
        \Phi(x) &= \mathtt{ELU}(x) + 1 \\
        \mathtt{ELU}(x) &=
        \begin{cases}
            x, & \text{if $x>$ 0} \\
            \exp(x) - 1, &\text{if $x\leq$ 0}
        \end{cases}
    \end{aligned}
    \label{eq:feature-function-elu}
\end{equation}
as proposed by \citet{linear-transformers-katharopoulos:2020}.

\subsection{Ablation Study}
We conduct an ablation study on the prominent parts of the proposed architecture, namely, the self-attention mechanism that allows the model to capture spatial dependencies and the conditioning mechanism that is used to condition the neural field. For 2D PDEs, we also study the effect on the model's performance for patches of one size and the multi-scale patching mechanism with small and large patches. Additionally, we compare linear attention \cite{linear-transformers-katharopoulos:2020} with vanilla attention \cite{transformers-vaswani} in terms of training time and GPU memory consumption. We perform the ablation study mainly on the 1D Burgers' and 2D CNS datasets for simplicity.

\paragraph{Self-Attention, Conditioning Mechanism, and Patch Generation.}
\cref{tab:ablation_study} shows the results of the proposed model with and without self-attention as well as with different modulation mechanisms to condition the neural field. \cref{tab:ablation_study_patches} presents the different results for patches of one size vs multi-scale patching mechanism.

\begin{table}[!htb]
    \small
    \centering
    \begin{tabular}{ cccll }
        \toprule
        PDE & Self-attention & Conditioning mechanism & nRMSE (\textbf{$\downarrow$}) & bRMSE (\textbf{$\downarrow$}) \\
        \midrule
        \multirow{3}{*}{1D Burgers} & \cmark & Modulation with scaling & \textbf{0.0824} & \textbf{0.0228} \\
        & \xmark & Modulation with scaling & 0.8890 & 0.3242 \\
        & \cmark & Modulation with scaling and shifting & 0.0945 & 0.0291 \\ 
        \bottomrule
    \end{tabular}
    \caption{Ablation study for attention mechanism and conditioning mechanism of our proposed VCNeF model.}
    \label{tab:ablation_study}
\end{table}

\begin{table}[!htb]
\small
\centering
    \begin{tabular}{ ccll }
        \toprule
        PDE & Patches & nRMSE (\textbf{$\downarrow$}) & bRMSE (\textbf{$\downarrow$}) \\
        \midrule
        \multirow{2}{*}{2D CNS} & Small and large & \textbf{0.1994} & \textbf{0.0904}  \\
        & Only large & 0.4569 & 0.1982 \\ 
        \bottomrule
    \end{tabular}
    \caption{Ablation study for the multi-scale mechanism of our proposed VCNeF model.}
    \label{tab:ablation_study_patches}
\end{table}

\paragraph{Vanilla Attention and Linear Attention.}
The Linear Transformer and the linear self-attention component in the proposed architecture can be replaced with vanilla attention or some arbitrary attention mechanism. We choose linear attention since it promises a speed-up for long sequences (i.e., fine resolution of the spatial domain) compared to vanilla attention. \cref{tab:vcnef-vanilla-linear-attention-comparison} shows empirical results for training the transformer-based VCNeF on the 1D Burgers' PDE. We observe that the memory of linear attention increases linearly and of vanilla attention quadratically. Double the spatial resolution corresponds to double the number of tokens, yielding an increased memory and time consumption. Training the VCNeF with vanilla attention requires more than 640 GiB, while the VCNeF with linear attention requires only 99.4 GiB. We use the vanilla attention implementation of \citet{linear-transformers-katharopoulos:2020} for a fair comparison to the non-optimized linear attention implementation.

\begin{table}[!htb]
    \begin{center}
        \small
        \begin{tabular}{ cccrc }
            \toprule
            PDE & Spatial resolution (\# tokens) & Attention type & GPU memory & Time per epoch \\
            \toprule
            \multirow{6}{*}{1D Burgers} & \multirow{2}{*}{256} & Vanilla & 72.6 GiB & 28 s \\
            & & Linear & 31.4 GiB & 18 s\\
            \cmidrule{2-5}
            & \multirow{2}{*}{512} & Vanilla & 223.4 GiB & 78 s\\
            & & Linear & 53.8 GiB & 32 s\\
            \cmidrule{2-5}
            & \multirow{2}{*}{1024} & Vanilla & $>$640 GiB & N/A \\
            & & Linear & 99.4 GiB & 62 s\\
            \bottomrule
        \end{tabular}
    \end{center}
    \caption{GPU memory consumption and training time per epoch for the VCNeF with vanilla attention (scaled dot-product attention) and linear attention on the 1D Burgers train set. The values refer to training with a batch size of 64 on 4x NVIDIA A100-SXM4 80GB GPUs using data parallelism. The number of queried timesteps $N_t$ is 40. Time per epoch includes the time that is needed to load the data and transfer it to the GPUs.}
    \label{tab:vcnef-vanilla-linear-attention-comparison}
\end{table}

\FloatBarrier
\section{PDE Dataset Details}
\label{app:dataset-details}

We conduct experiments on the following four challenging hydrodynamical equations of time-dependent parametric PDE datasets from PDEBench~\cite{pdebench-takamoto:2022}.

\subsection{1D Burgers' Equation}
The Burgers' PDE models the non-linear behavior and diffusion process in fluid dynamics and is expressed as
\begin{equation}
    \partial_t u(t,x) + u(t,x) \partial_x u(t,x) = \frac{\nu}{\pi}  \partial_{xx}u(t,x)
    \label{eq:burgers-pde}
\end{equation}
where the PDE parameter $\nu$ denotes the diffusion coefficient. Our dataset contains solutions for x $\in$ (-1, 1) with a maximum resolution of 1024 spatial discretization points and t $\in$ (0, 2] with a maximum resolution of 201 temporal discretization steps, including the initial condition. We subsample the data along the temporal and spatial domain, yielding a trajectory of 41 timesteps, with each snapshot having a spatial resolution of 256.

\subsection{1D Advection Equation}
The Advection PDE models pure advection behavior without non-linearity. It is written as
\begin{equation}
    \partial_t u(t,x) + {\beta} \partial_x u(t,x) = 0
    \label{eq:advection-pde}
\end{equation}
where the PDE parameter $\beta$ denotes the advection velocity. Similar to 1D Burgers, we subsample the data to get a trajectory of 41 time steps, each with a spatial resolution of 256.

\subsection{1D, 2D, and 3D Compressible Navier-Stokes (CNS) Equations}

The Navier-Stokes equations (Equations \ref{eq:comp-nast-1} to \ref{eq:comp-nast-3}) are a compressible version of fluid dynamics equations that describe the flow of a fluid. Thus, the equations are important for Computational Fluid Dynamics (CFD) applications. Equation \ref{eq:comp-nast-1} refers to the mass continuity equation, which is also called the transport equation or equation of conservation of mass, Equation \ref{eq:comp-nast-2} describes the conservation of momentum, and Equation~\ref{eq:comp-nast-3} represents energy conservation.
\begin{subequations}\label{eq:2d-comp-nast}
    \begin{align} 
        \partial_t \rho + \nabla \cdot ({\rho} \textbf{v}) &= 0, \label{eq:comp-nast-1} \\
        \rho (\partial_t \textbf{v} + \textbf{v} \cdot \nabla \textbf{v}) &= - \nabla p + \eta \triangle \textbf{v} + ({\zeta} + \frac{\eta}{3}) \nabla (\nabla \cdot \textbf{v}),
        \label{eq:comp-nast-2}\\
        \partial_t (\epsilon + \frac{\rho v^2}{2}) &+ \nabla \cdot [( p + {\epsilon} + \frac{\rho v^2}{2} ) \bf{v} - \bf{v} \cdot {\sigma'} ] = 0,\label{eq:comp-nast-3}
    \end{align}
\end{subequations}
where $\rho$ represents the (mass) density of the fluid, $\bf{v}$ denotes the fluid velocity (in vector field), p stands for the gas pressure, $\epsilon$ describes the internal energy according to the equation of state, $\sigma'$ is the viscous stress tensor, $\eta$ and $\zeta$ are the PDE parameters which represent the shear and bulk viscosity, respectively. We subsample the data for the 1D, 2D, and 3D equations. The original resolution of the 1D simulation data has 1024 spatial points and 101 timesteps. For training purposes, we subsample across both spatial and temporal resolutions by a factor of 4 and 2, respectively, yielding a trajectory of length 51 time steps and a spatial resolution of 256. To be consistent with other 1D PDE trajectory lengths, we retain only the first 41 timesteps and perform experiments on this truncated data. The original resolution of 2D simulation is $128\times128$ for each channel (i.e., density, velocity-x, velocity-y, and pressure) and has 21 timesteps. For training, we have subsampled the data only for the spatial dimension, resulting in a resolution of $64\times64$. For 3D data, the original spatial resolution is $128 \times 128 \times 128$, which was subsampled to a resolution of $32 \times 32 \times 32$ for training. The temporal resolution is 21 timesteps for the 2D case.

Table~\ref{tab:comp-ns-2d-dataset-config} summarizes the spatial and temporal discretization, Mach number, and PDE parameter values of the 2D CNS dataset.

\begin{table}[!htbp]
    \begin{center}
        \small
        \begin{tabular}{ ccccccc }
            \toprule
            \makecell{Field Type} &\makecell{Mach}  & \makecell{$\eta$} & \makecell{$\zeta$}  &\makecell{$\Delta$t} &\makecell{$\Delta$x} &\makecell{$\Delta$y}\\
            \midrule
            Rand &\{0.1, 1.0\}  & 0.01 & 0.01 & 0.05 & 0.0078125 & 0.0078125\\
            Rand &\{0.1, 1.0\} & 0.1  & 0.1  & 0.05 & 0.0078125 & 0.0078125\\
            \midrule
            Rand &\{0.1, 1.0\} & $10^{-8}$ & $10^{-8}$  & 0.05 & 0.001953125 & 0.001953125\\
            \bottomrule
        \end{tabular}
    \end{center}
    \caption{Original dataset configuration for 2D Compressible Navier-Stokes equations.}
    \label{tab:comp-ns-2d-dataset-config}
\end{table}

However, since we perform subsampling on the spatial axes before training the neural net models, the effective values of $\Delta$x and $\Delta$y are slightly higher (i.e., coarser resolution).

\section{Baseline Models}
\begin{figure}[!htbp]
    \begin{center}
        \includegraphics[width=0.88\textwidth]{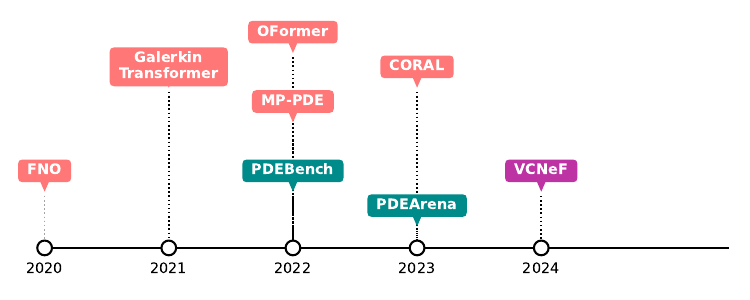}
        \caption{Timeline of Neural PDE Solvers and Benchmarks (PDEBench \& PDEArena). Proposed model \textcolor{mediumorchid}{VCNeF}.}
        \label{fig:timeline-neural-pde-solvers}
    \end{center}
\end{figure}

We consider the following strong baselines spanning five different families of models for solving PDEs using neural networks: Fourier Neural Operators, GNNs, UNets, Neural Fields,  and Transformers. Our proposed model is indicated in pink color in Figure~\ref{fig:timeline-neural-pde-solvers}.

\subsection{Fourier Neural Operator Baseline}
\label{app:baselines-fno}

Fourier Neural Operator (FNO) \cite{fno-li} is an implementation of a neural operator that maps from one function $a(x)$ to another function $a'(x)$~\cite{neural-operators-kovachki}. Traditionally, neural networks model a mapping between two finite-dimensional Euclidean spaces which leads to the problem that they are fixed to a spatial and temporal resolution when used for solving PDEs. Neural operators overcome this limitation by learning an operator that is a mapping between infinite-dimensional function spaces (i.e., mapping between functions). FNO, an instantiation of a neural operator, is based on spectral convolution layers that implement an integral transformation of the input function. The integral transformation is implemented with discrete Fourier transforms on the spatial or spatial and temporal domain, allowing for efficient and expressive architecture. We use an FNO model that applies the integral transformation on the spatial domain and is trained in an autoregressive fashion for 500 epochs as a baseline. Thus, the FNO learns a mapping between the function $a(\boldsymbol{x}) := u(t_n, \boldsymbol{x})$ representing the solution for timestep $t_n$ and $a'(\boldsymbol{x}) := u(t_{n+1}, \boldsymbol{x})$ denoting the solution for a future timestep $t_{n+1}$.

\paragraph{cFNO.}
cFNO \citep{cape-takamoto:2023} is the adapted version of the FNO where the PDE parameters are added as an additional channel to condition the model on the PDE parameters.

\subsection{Graph Neural Network Baseline}
\label{app:baselines-gnn}
Several models exist that use Graph Neural Networks (GNNs) for solving PDEs~\cite{mp-pde-brandstetter,magnet-pde-boussif:2022}

\paragraph{Message Passing Neural PDE Solvers.}
MP-PDE~\cite{mp-pde-brandstetter} follows the prevalent \emph{Encode-Process-Decode} framework for simulating physical systems~\cite{learn2simulate-physics-sanchez-gonzalez:2020}. The MP-PDE model has an MLP as an encoder, GNN as a processor, and a CNN as the decoder. Moreover, the model introduces several tricks such as pushforward, temporal bundling (time window with 5 timesteps), and random timesteps in the length of the trajectory as starting points during training for autoregressive PDE solving, while also considering the PDE parameter values as additional input, making it as a versatile choice for generalized neural PDE solving. Hence, we adopt it as a baseline. However, it has to be noted that we apply the model only to 1D PDEs. The configuration of our adaptation for 1D PDEs amounts to 614,929 model parameters, which is comparable to the other baselines.

\subsection{UNet Baselines}
\label{app:baselines-unet}
UNet-style models are increasingly becoming popular choices for weather forecasting and PDE solving due to their natural support for multi-scale data modeling~\cite{pdebench-takamoto:2022,unet-neural-op-uno-rahman:2023,ditto-diffusion-temporal-transformer-ovadia:2023,pde-refiner-long-term-rollouts-neural-pde-lippe:2023}, and have, thus, emerged as one of the competitive baseline models in SciML literature.

\paragraph{UNet-PDEArena.}
\label{app:baselines-unet-pdearena}
We use the modern UNet from PDEArena\footnote{\url{https://github.com/pdearena/pdearena/}}~\cite{generalized-pde-modeling-gupta:2023}. To match the number of parameters of the other models, we use an initial hidden dimension of 16 and two downsampling layers (3 for 2D). Following \citet{pde-refiner-long-term-rollouts-neural-pde-lippe:2023}, we use the model to predict the residual instead of the next step directly and scale down the model output by a factor of $0.3$. The training is performed autoregressively for 500 epochs using an initial learning rate of $3.e^{-3}$, which gets halved every 100 epochs.

\subsection{Implicit Neural Representation Baseline}
\label{app:baselines-inr-coral}

\paragraph{CORAL: Coordinate-based Model for Operator Learning.} Considering that CORAL~\cite{operator-learning-neural-fields-general-geometries-serrano:2023}, to the best of our knowledge, is the current state-of-the-art INR-based method for solving PDEs, we benchmark our proposed VCNeF model against it on 1D Advection and Burgers' as well as on the challenging 1D and 2D compressible Navier-Stokes PDEs~\cite{pdebench-takamoto:2022}. The CORAL model is trained purely in a data-driven manner and involves two stages: (i) INR training and (ii) Dynamics Modeling training. Due to the sequential nature of this two-phase training, we first train the INR model, and the dynamics model is trained after the completion of the INR model training. CORAL authors conducted experiments on a small dataset of 256 training and 16 test samples. We, on the other hand, conduct experiments on PDEBench which consists of 9000 train and 1000 test samples. Hence, we train the CORAL baseline model for 1000 epochs for INR training and 500 epochs for dynamics modeling optimization, unlike the original authors' suggested setting of 10000 epochs of optimization for both INR and dynamics modeling training.

As with other baseline models, we train and test the CORAL baseline on the subsampled data, yielding a spatial and temporal resolution of 256 and 41, respectively. We report results on 1D Advection, Burgers, and CNS. The training of 2D CNS resulted in very high errors in the INR training phase and the loss diverged to NaN values in dynamics modeling training. We encode both the single and multiple channel inputs of PDEs in a single latent space of dimension 256 with the aim to keep the model simple and match the number of parameters to other baseline models. For other hyperparameter values such as the learning rate, NODE depth, and width, we use the default values suggested by \citet{operator-learning-neural-fields-general-geometries-serrano:2023}.

\subsection{Transformer Baselines}
\label{app:baselines-transformers}
Transformer-based models are increasingly used to solve PDEs~\cite{galerkin-cao,oformer-pde-li:2023,scalable-transformer-pde-li:2023,gnot-operator-learn-hao:2023}. Hence, we use Galerkin Transformer and Operator Transformer as state-of-the-art transformer-based models.

\paragraph{Galerkin Transformer.}
\citet{galerkin-cao} introduces the novel application of self-attention for learning a neural operator. The author provides an alternative way to interpret the matrices $\boldsymbol{Q}, \boldsymbol{K}, \boldsymbol{V}$ by interpreting them column-wise as the evaluation of learned basis functions instead of row-wise as the latent representation of the tokens. This new interpretation allows the author to improve the effectiveness of the attention mechanism by linearizing it, yielding Fourier and Galerkin-type attention. The author employs the proposed attention mechanisms in a transformer-based neural operator for solving PDEs. We choose the Galerkin Transformer as a baseline because it is transformer-based and uses self-attention on the spatial domain of the PDE. The baseline model is trained for 500 epochs in an autoregressive fashion using the hyperparameters suggested by \citet{galerkin-cao}.
\paragraph{Operator Transformer.}
OFormer~\cite{oformer-pde-li:2023} is a transformer-based neural operator that is based on the attention types proposed in Galerkin and Fourier Transformer \citep{galerkin-cao}. Existing approaches such as FNO and Galerkin or Fourier Transformer are restricted in having the same grid for the input and output. Consequently, it is impossible to query the model (i.e., output function) on arbitrary spatial points that differ or partially disjoint from the input points. OFormer solves this problem by adding cross-attention to the model to allow querying for arbitrary spatial points. In addition, the authors suggest further improvements to the Galerkin or Fourier Transformer and name the resulting model Operator Transformer (OFormer). We train the OFormer model in an autoregressive fashion with the curriculum learning strategy of \citet{cape-takamoto:2023} for 500 epochs.
\paragraph{cOFormer.}
Inspired by cFNO \citep{cape-takamoto:2023}, we adapt OFormer to take the PDE parameter as an additional input. The PDE parameter values are appended to the input as an additional channel to condition the model on the PDE parameter.

\section{Additional Experimental Details}

\subsection{Used PDE Parameters}
\cref{tab:pde_params_experiments_multiple} shows the combinations of PDE parameter values used in our experiments for the multiple parameters setting. In this case, we train the models on a set of PDE parameter values (\textbf{seen}) and test it on a different set of PDE parameter values (\textbf{unseen}) with the aim of testing the model's generalization capabilities on this aspect.

\begin{table}[!htb]
    \centering
    \begin{tabular}{ ccc }
        \toprule
        PDE & Training Set Parameters  (seen) & Test Set Parameters (unseen) \\
        \midrule
        1D Burgers & $\nu = (0.002, 0.004, 0.02, 0.04, 0.2, 0.4, 2.0)$ & $\nu = (0.001, 0.01, 0.1, 1.0, 4.0)$ \\
        1D Advection & $\beta = (0.2, 0.4, 0.7, 2.0, 4.0)$ & $\beta = (0.1, 1.0, 7.0)$ \\
        1D CNS & $\eta = \zeta =(10^{-8}, 0.001, 0.004, 0.01, 0.04, 0.1)$ & $\eta = \zeta = (0.007, 0.07)$ \\
        \bottomrule
    \end{tabular}
    \caption{Exemplary set of PDE parameters used in our experiments with multiple PDE parameters.}
    \label{tab:pde_params_experiments_multiple}
\end{table}

\subsection{Loss Function}
We use the Mean Squared Error (MSE) loss function as the optimization criterion for FNO, UNet, Galerkin Transformer, OFormer, and the proposed VCNeF. Let $\boldsymbol{Y} \in \mathbb{R}^{N_b \times N_t \times s \times c}$ be a batch of ground truth trajectories and $\boldsymbol{\hat{Y}} \in \mathbb{R}^{N_b \times N_t \times s \times c}$ the corresponding batch of model's predictions where $N_b$ denotes the batch size, $N_t$ is the length of the trajectories, $s = s_x \cdot s_y \cdot \hdots$~the spatial points per timestep, and $c$ the number of channels of the PDE. Then, the MSE loss function is defined as

\begin{equation}
    \text{MSE}(\boldsymbol{\hat{Y}}, \boldsymbol{Y}) = \frac{1}{N_b \cdot N_t \cdot s \cdot c} \sum_{b=1}^{N_b} \sum_{t=1}^{N_t} \sum_{i=1}^{s} \sum_{j=1}^c (\boldsymbol{\hat{Y}}_{b, t, i, j} - \boldsymbol{Y}_{b, t, i, j})^2
    \label{eq:loss_function_mse}
\end{equation}

MP-PDE and CORAL are trained on the loss functions suggested by the authors.

\subsection{Model's Hyperparameters}
Tables \ref{tab:fno-model-hyperparams-single}, \ref{tab:mp-pde-model-hyperparams-single}, \ref{tab:transformer-model-hyperparams-single}, \ref{tab:unet-pdebench-pdearena-model-hyperparams-single} list the used hyperparameters for the baselines and our proposed VCNeF.

\begin{sidewaystable}
    \centering
    \tiny
    \begin{tabular}{ cccccccccccc }
        \toprule
         PDE & Model & Epochs & Batch size & Fourier width & \# Fourier Modes & \# Layers & Learning  rate & LR Scheduler & \# Parameters & Curriculum learning \\
        \midrule
        1D Burgers & FNO & 500 & 64 & 64 & 16 & 4 & 1.e-4 & Step Scheduler & 549,569 & \xmark \\
        \midrule
        1D Advection & FNO & 500 & 64 & 64 & 16 & 4 & 1.e-4 & Step Scheduler & 549,569 & \xmark \\
        \midrule
        1D CNS & FNO & 500 & 64 & 64 & 16 & 4 & 6.e-5 & Step Scheduler & 549,955 & \xmark \\
        \midrule
        2D CNS & FNO & 500 & 64 & 32 & 12 & 4 & 3.e-4 & Step Scheduler & 9,453,716 & \xmark \\
        \midrule
        3D CNS & FNO & 1000 & 4 & 20 & 12 & 4 & 3.e-4 & Step Scheduler & 22,123,753 & \xmark \\
        \bottomrule
    \end{tabular}
    \caption{Hyperparameters for the FNO used in the single PDE parameter experiments. The Step Scheduler was configured with a step size of 100 and a gamma of 0.5.}
    \label{tab:fno-model-hyperparams-single}

    \bigskip

    \centering
    \tiny
    \begin{tabular}{ cccccccccc }
        \toprule
         PDE & Model & Epochs & Batch size & Embedding size & \# Layers & Time window & Learning rate  & \# Parameters & Curriculum learning \\
        \midrule
        1D Burgers & MP-PDE & 20 & 64 & 128 & 6 & 5 & 1.e-4 & 614,929 & \xmark \\
        \midrule
        1D Advection & MP-PDE & 20 & 64 & 128 & 6 & 5 & 1.e-4 & 614,929 & \xmark \\
        \bottomrule
    \end{tabular}
    \caption{Hyperparameters for the MP-PDE used in the single PDE parameter experiments.}
    \label{tab:mp-pde-model-hyperparams-single}

    \bigskip
    
    \centering
    \tiny
    \begin{tabular}{ cccccccccccc }
        \toprule
         PDE & Model & Epochs & Batch size & Embedding  size & \# Heads & \# Layers & Learning rate & LR Scheduler & \# Parameters & Curriculum learning \\
        \midrule
        \multirow{3}{*}{1D Burgers} & OFormer & 500 & 64 & 96 & 1 & 4+3 & 6.e-5 & One Cycle Scheduler & 660,814 & \cmark \\
        & Galerkin Transformer & 500 & 64 & 96 & 1 & 4+2 & 1.e-5 & One Cycle Scheduler & 530,305 & \xmark \\
        & VCNeF & 500 & 32 & 96 & 8 & 3+3 & 3.e-4 & One Cycle Scheduler & 793,825 & \xmark \\
        \midrule
        \multirow{3}{*}{1D Advection} & OFormer & 500 & 64 & 96 & 1 & 4+3 & 6.e-5 & One Cycle Scheduler & 660,814 & \cmark \\
        & Galerkin Transformer & 500 & 64 & 96 & 1 & 4+2 & 1.e-5 & One Cycle Scheduler & 530,305 & \xmark \\
        & VCNeF & 500 & 64 & 96 & 8 & 3+3 & 6.e-4 & One Cycle Scheduler & 793,825 & \xmark \\
        \midrule
        \multirow{3}{*}{1D CNS} & OFormer & 500 & 64 & 96 & 1 & 4+3 & 6.e-5 & One Cycle Scheduler & 662,733 & \cmark \\
        & Galerkin Transformer & 500 & 64 & 96 & 1 & 4+2 & 1.e-5 & One Cycle Scheduler & 530,595 & \xmark \\
        & VCNeF & 500 & 64 & 96 & 8 & 3+3 & 4.e-4 & One Cycle Scheduler & 794,307 & \xmark \\
        \midrule
        \multirow{2}{*}{2D CNS} & Galerkin Transformer & 500 & 64 & 384 & 1 & 6+2 & 8.e-5 & One Cycle Scheduler & 8,053,091 & \xmark \\
        & VCNeF & 1000 & 64 & 256 & 8 & 1+6 & 3.e-4 & One Cycle Scheduler & 11,779,436 & \xmark \\
        \midrule
        3D CNS & VCNeF & 1000 & 4 & 256 & 8 & 1+6 & 3.e-4 & One Cycle Scheduler & 27,335,041 & \xmark \\
        \bottomrule
    \end{tabular}
    \caption{Hyperparameters for the transformer-based models used in the single PDE parameter experiments. The One Cycle Scheduler was configured to reach the maximum learning rate at 0.2, start division factor 1.e-3 and final division factor 1.e-4.}
    \label{tab:transformer-model-hyperparams-single}

    \bigskip
     
    \centering
    \tiny
    \begin{tabular}{ cccccccccccc }
        \toprule
         PDE & Model & Epochs & Batch size & \# Downsample & \# Upsample & \# Layers & Learning  rate & LR Scheduler & \# Parameters & Curriculum learning \\
        \midrule
        1D Burgers & UNet  & 500 & 256 & 64 & 16 & 4 & 1.e-4 & Step Scheduler & 557,137 & \xmark \\
        \midrule
        1D Advection & UNet & 500 & 256 & 64 & 16 & 4 & 1.e-4 & Step Scheduler & 557,137 & \xmark \\
        \midrule
        1D CNS & UNet & 500 & 256 & - & 16 & 4 & 6.e-5 & Step Scheduler & 562,579 & \xmark \\
        \midrule
        2D CNS & UNet & 500 & 256 & 246 & 12 & 4 & 3.e-4 & Step Scheduler & 9,187,284 & \xmark \\
        \bottomrule
    \end{tabular}
    \caption{Hyperparameters of UNets from PDEArena for single PDE parameter experiments. The Step Scheduler was configured with a step size of 80 and a gamma of 0.5.}
    \label{tab:unet-pdebench-pdearena-model-hyperparams-single}
\end{sidewaystable}

\FloatBarrier
\subsection{Evaluation Metrics}
We use the normalized RMSE (nRSME) and boundary RMSE (bRMSE) from PDEBench \citep{pdebench-takamoto:2022} as metrics to evaluate the models.
\paragraph{Normalized RMSE (nRMSE).} The normalized RMSE ensures the independence of the different scales of field variables. The channels of PDEs with multiple channels are often on different scales (e.g., one channel consists of values with small magnitudes while another channel consists of values with large magnitudes). Additionally, the scale of a single channel usually changes when the time-dependent PDE evolves in time (e.g., large magnitudes at the beginning of the trajectory decaying to small magnitudes at the end). nRMSE is independent of these scaling effects and provides a good metric for the global and local performance of the ML model. Let $\boldsymbol{Y} \in \mathbb{R}^{N_t \times s \times c}$ be the ground truth trajectory and $\boldsymbol{\hat{Y}} \in \mathbb{R}^{N_t \times s \times c}$ the model's prediction where $N_t$ denotes the length, $s = s_x \cdot s_y \cdot \hdots$~the spatial points per timestep, and $c$ the number of channels of the PDE. Then, the per-sample nRMSE is defined as
\begin{equation}
    \begin{aligned}
        \text{relativeError}(t, c') &= \frac{ \norm{\boldsymbol{Y}_{t, \cdot, c'} - \boldsymbol{\hat{Y}}_{t, \cdot, c'}}_2 } { \norm{\boldsymbol{Y}_{t, \cdot, c'}}_2 } \qquad \in \mathbb{R} \\
        \text{nRMSE} &= \frac{1}{N_t \cdot c} \sum_{t=1}^{N_t} \sum_{i=1}^c \text{relativeError}(t, i) \qquad \in \mathbb{R}
    \end{aligned}
    \label{eq:normalized_error}
\end{equation}

\paragraph{Boundary RMSE (bRMSE).} The RMSE on the boundaries of the spatial domain quantifies whether the boundary condition can be learned or not. Let $\boldsymbol{Y} \in \mathbb{R}^{N_t \times s_x \times c}$ be the ground truth trajectory of a 1D PDE and $\boldsymbol{\hat{Y}} \in \mathbb{R}^{N_t \times s_x \times c}$ the model's prediction where $N_t$ denotes the length, $s_x$ denotes the number of points for the x-axis, and $c$ the number of channels of the PDE under consideration. Then, the per-sample bRMSE is defined as
\begin{equation}
    \begin{aligned}
        \text{boundaryError}(t, c') &= \sqrt{\frac{ (\boldsymbol{Y}_{t, 1, c'} - \boldsymbol{\hat{Y}}_{t, 1, c'} )^2 + (\boldsymbol{Y}_{t, s, c'} - \boldsymbol{\hat{Y}}_{t, s, c'} )^2}{2}} \qquad \in \mathbb{R} \\
        \text{bRMSE} &= \frac{1}{N_t \cdot c} \sum_{t=1}^{N_t} \sum_{i=1}^c \text{boundaryError}(t, i) \qquad \in \mathbb{R}
    \end{aligned}
    \label{eq:boundary_error}
\end{equation}

\subsection{Randomized Starting Points Training}
Algorithm \ref{alg:random-points} describes the training with randomized starting points for the VCNeF-R. During training, the model is conditioned on the initial value $u(0, \boldsymbol{x})$ as well as on randomly sampled $u(t, \boldsymbol{x})$ with $t \sim \mathcal{U}\{0, T\}$ along the trajectory as starting points.

\begin{algorithm}[!htb]
   \caption{Randomized Starting Points as Initial Conditions}
   \label{alg:random-points}
    \begin{algorithmic}
        \STATE {\bfseries Input:} training data set $\mathcal{D}$, training trajectories length $N_t$, number of epochs $Epochs$
        \FOR{epoch = 1 {\bfseries to} $Epochs$}
            \STATE starting\_points = [0] $\cup$ random\_shuffle([1, $\cdots$, $N_t-1$])[:10]
            \FOR{starting\_point {\bfseries in} starting\_points}
                \FOR{(x, y) {\bfseries in} $\mathcal{D}$}
                    \STATE model.train(x[starting\_point], y[starting\_point:])
                \ENDFOR
            \ENDFOR
       \ENDFOR
    \end{algorithmic}
\end{algorithm}

\FloatBarrier
\section{Additional Experimental Results}
\label{app:results}

This section contains additional results of the experiments. We train all models on two different initializations and provide the mean and standard deviations of the runs. Similar to the experiments section in the main paper, we structure the results to answer the following five research questions.

\begin{description}
    \item[Q1:] How effective are VCNeFs compared to the state-of-the-art (SOTA) methods when trained and tested for the same PDE parameter value?
    \item[Q2:] How well can VCNeFs generalize to PDE parameter values not seen during training?
    \item[Q3:] How well can VCNeFs do spatial and temporal zero-shot super-resolution?
    \item[Q4:] Does training on initial conditions sampled from training trajectories improve the accuracy?
    \item[Q5:] Does the vectorization provide a speed-up, and what is the model's scaling behavior?
\end{description}

\subsection{(Q1): Comparison to state-of-the-art baselines}

\subsubsection{Detailed Metrics}
The Tables \ref{tab:burgers}, \ref{tab:advection}, \ref{tab:1d-cfd}, \ref{tab:2d-cfd}, and \ref{tab:3d-cfd} show the metrics with standard deviations for the chosen PDEs and models.

\begin{table}[!htb]
    \begin{minipage}{.45\linewidth}
        \begin{center}
            \begin{tabular}{ lll }
                \toprule
                Model & nRMSE (\textbf{$\downarrow$}) & bRMSE (\textbf{$\downarrow$}) \\
                \midrule
                FNO & 0.0987$^{\pm0.0004}$ & 0.0225$^{\pm0.0006}$ \\
                MP-PDE & 0.3046$^{\pm0.0004}$ & 0.0725$^{\pm0.0014}$ \\
                UNet & \textbf{0.0566}$^{\pm0.0004}$ & 0.0259$^{\pm0.0019}$ \\
                CORAL & 0.2221$^{\pm0.0108}$ & 0.0515$^{\pm0.0001}$ \\
                Galerkin & 0.1651$^{\pm0.0044}$ & 0.0366$^{\pm0.0012}$ \\
                OFormer & 0.1035$^{\pm0.0059}$ & 0.0215$^{\pm0.0009}$ \\
                VCNeF & 0.0824$^{\pm0.0004}$ & 0.0228$^{\pm0.0003}$ \\
                VCNeF-R & 0.0784$^{\pm0.0001}$ & \textbf{0.0179}$^{\pm0.0001}$ \\
                \bottomrule
            \end{tabular}
            \caption{Normalized RMSE (nRMSE) and RMSE at the boundaries (bRMSE) of baselines and proposed model for the 1D Burgers' equation with $\nu = 0.001$.}
            \label{tab:burgers}
        \end{center}
    \end{minipage}%
    \hfill
    \begin{minipage}{.45\linewidth}
        \begin{center}
            \begin{tabular}{ lll }
                \toprule
                Model & nRMSE (\textbf{$\downarrow$}) & bRMSE (\textbf{$\downarrow$}) \\
                \midrule
                FNO & 0.0190$^{\pm0.0003}$ & 0.0239$^{\pm0.0002}$ \\
                MP-PDE & 0.0195$^{\pm0.0011}$ & 0.0283$^{\pm0.0022}$ \\
                UNet & \textbf{0.0079}$^{\pm0.0024}$ & 0.0129$^{\pm0.0043}$ \\
                CORAL & 0.0198$^{\pm0.0031}$ & 0.0127$^{\pm0.0014}$ \\
                Galerkin & 0.0621$^{\pm0.0024}$ & 0.0349$^{\pm0.0011}$ \\
                OFormer & 0.0118$^{\pm0.0012}$ & 0.0073$^{\pm0.0008}$ \\
                VCNeF & 0.0165$^{\pm0.0007}$ & 0.0088$^{\pm0.0003}$ \\
                VCNeF-R & 0.0113$^{\pm0.0003}$ & \textbf{0.0040}$^{\pm0.0005}$ \\
                \bottomrule
            \end{tabular}
            \caption{Normalized RMSE (nRMSE) and RMSE at the boundaries (bRMSE) of baselines and proposed model for the 1D Advection equation with $\beta = 0.1$.}
            \label{tab:advection}
        \end{center}
    \end{minipage} 
\end{table}

\begin{table}[!htb]
    \begin{minipage}{.45\linewidth}
        \begin{center}
            \begin{tabular}{ lll }
                \toprule
                Model & nRMSE (\textbf{$\downarrow$}) & bRMSE (\textbf{$\downarrow$}) \\
                \midrule
                FNO & 0.5722$^{\pm0.0244}$ & 1.9797$^{\pm0.0029}$ \\
                UNet & 0.2270$^{\pm0.0133}$ & \textbf{1.0399}$^{\pm0.0863}$ \\
                CORAL & 0.5993$^{\pm0.1014}$ & 1.5908$^{\pm0.1341}$ \\
                Galerkin & 0.7019$^{\pm0.0002}$ & 3.0143$^{\pm0.0112}$ \\
                OFormer & 0.4415$^{\pm0.0115}$ & 2.0478$^{\pm0.0581}$ \\
                VCNeF & 0.2943$^{\pm0.0034}$ & 1.3496$^{\pm0.0254}$ \\
                VCNeF-R & \textbf{0.2029}$^{\pm0.0227}$ & 1.1366$^{\pm0.0589}$ \\
                \bottomrule
            \end{tabular}
            \caption{Normalized RMSE (nRMSE) and RMSE at the boundaries (bRMSE) of baselines and proposed model for the 1D CNS equation with $\eta = \zeta = 0.007$.}
            \label{tab:1d-cfd}
        \end{center}
    \end{minipage}%
    \hfill
    \begin{minipage}{.45\linewidth}
        \begin{center}
            \begin{tabular}{ lll }
                \toprule
                Model & nRMSE (\textbf{$\downarrow$}) & bRMSE (\textbf{$\downarrow$}) \\
                \midrule
                FNO & 0.5625$^{\pm0.0015}$ & 0.2332$^{\pm0.0001}$ \\
                UNet & 1.4240$^{\pm0.5018}$ & 0.3703$^{\pm0.0432}$ \\
                Galerkin & 0.6702$^{\pm0.0036}$ & 0.8219$^{\pm0.0043}$ \\
                VCNeF & \textbf{0.1994}$^{\pm0.0086}$ & \textbf{0.0904}$^{\pm0.0036}$ \\
                \bottomrule
            \end{tabular}
            \caption{Normalized RMSE (nRMSE) and RMSE at the boundaries (bRMSE) of baselines and proposed model for the 2D CNS equation with $\eta = \zeta = 0.01$.}
            \label{tab:2d-cfd}
        \end{center}
    \end{minipage} 
\end{table}

\begin{table}[!htb]
    \begin{center}
        \begin{tabular}{ lll }
            \toprule
            Model & nRMSE (\textbf{$\downarrow$}) & bRMSE (\textbf{$\downarrow$}) \\
            \midrule
            FNO & 0.8138$^{\pm0.0007}$ & 6.0407$^{\pm0.0493}$ \\
            VCNeF & \textbf{0.7086}$^{\pm0.0005}$ & \textbf{4.8922}$^{\pm0.0077}$ \\
            \bottomrule
        \end{tabular}
        \caption{Normalized RMSE (nRMSE) and RMSE at the boundaries (bRMSE) of baselines and proposed model for the 3D CNS equation with $\eta = \zeta = 10^{-8}$.}
        \label{tab:3d-cfd}
    \end{center}
\end{table}

\FloatBarrier
\subsubsection{Error Visualization}
We visualize the error along the spatial and temporal domains for 1D PDEs using a heatmap. For a given timestep $t$, let $\boldsymbol{y_t} \in \mathbb{R}^{s}$ be the ground truth and $\boldsymbol{\hat{y_t}} \in \mathbb{R}^{s}$ the model's prediction. Then, we calculate the point-wise error for the 2D error visualization as

\begin{equation}
    e(t) = \sqrt{\frac{(\boldsymbol{y_t}-\boldsymbol{\hat{y_t}})^2}{\boldsymbol{y_t}^2}} = \frac{|\boldsymbol{y_t}-\boldsymbol{\hat{y_t}}|}{|\boldsymbol{y_t}|} \in \mathbb{R}^{s}
    \label{eq:rel-error-1d}
\end{equation}

\begin{figure}[!htb]
    \begin{center}
        \includegraphics[width=0.75\textwidth]{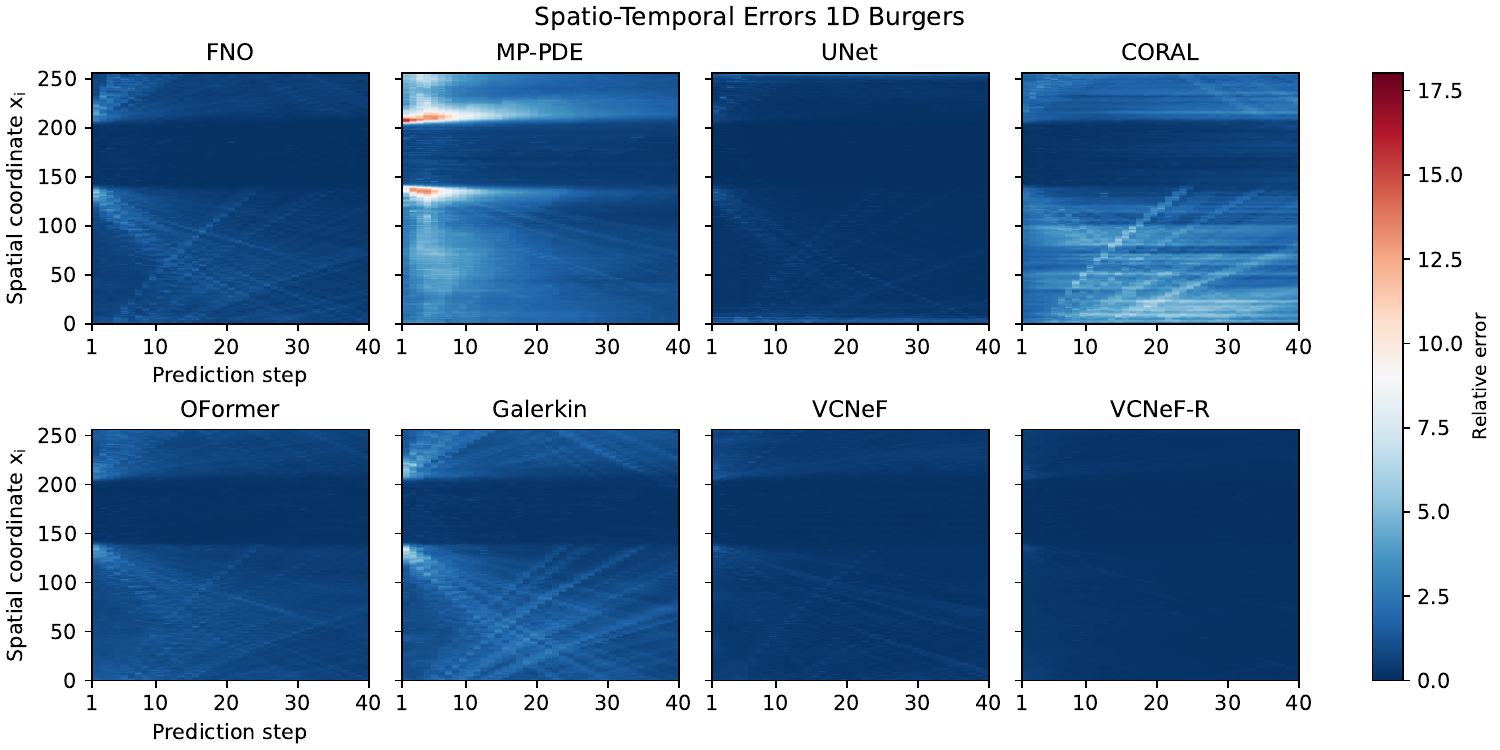}
        \caption{Error heatmap for 1D Burger's equation with $\nu = 0.001$. Models are trained and tested on spatial resolution $s = 256$ and temporal resolution $N_t = 41$. $t_0$ (not depicted above) is the initial condition.}
        \label{fig:burgers-heatmap}
    \end{center}
\end{figure}

\begin{figure}[!htb]
    \begin{center}
        \includegraphics[width=0.75\textwidth]{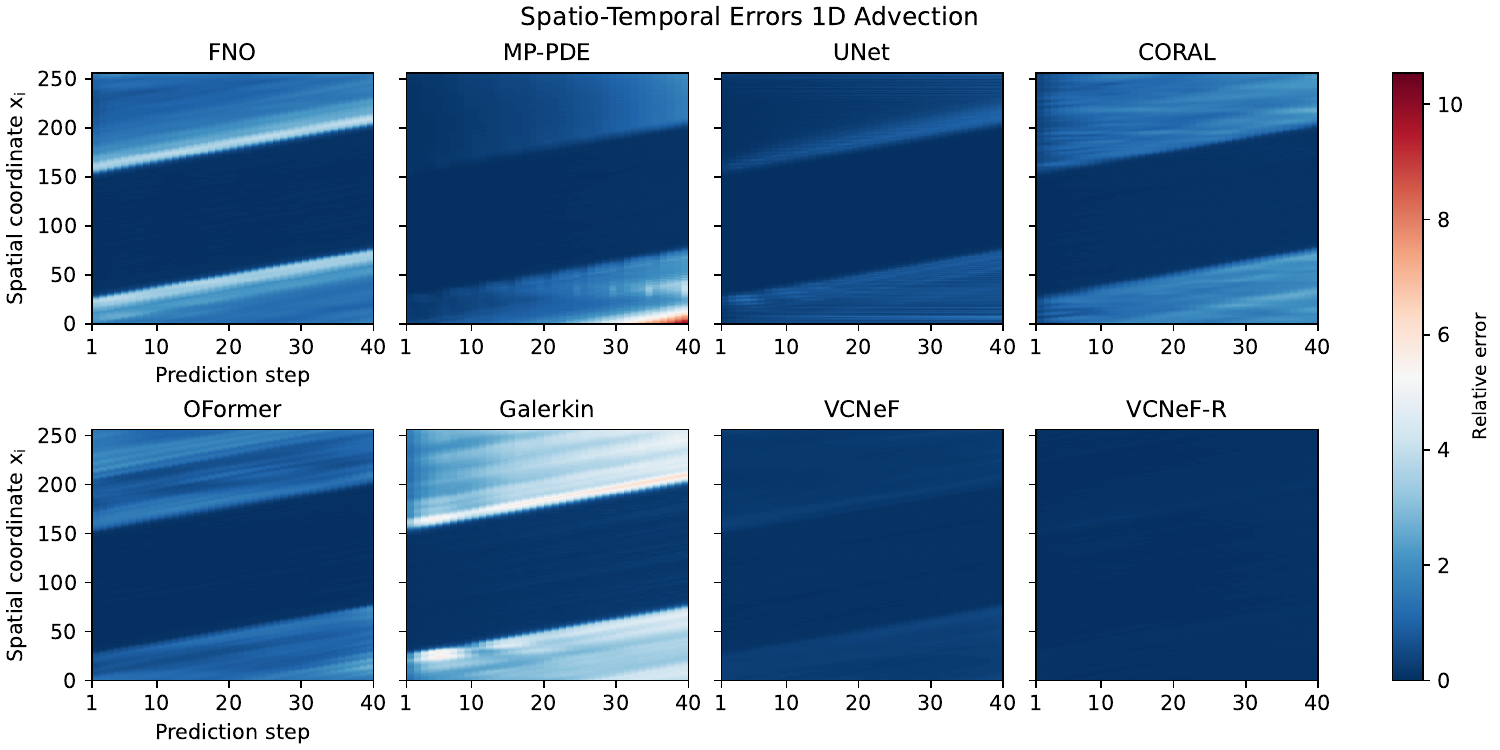}
        \caption{Error heatmap for 1D Advection with $\beta = 0.1$. Models are trained and evaluated on spatial resolution $s = 256$ and temporal resolution $N_t = 41$. $t_0$ (not shown above) is the initial condition.}
        \label{fig:advection-heatmap}
    \end{center}
\end{figure}

\begin{figure}[!htb]
    \begin{center}
        \includegraphics[width=0.75\textwidth]{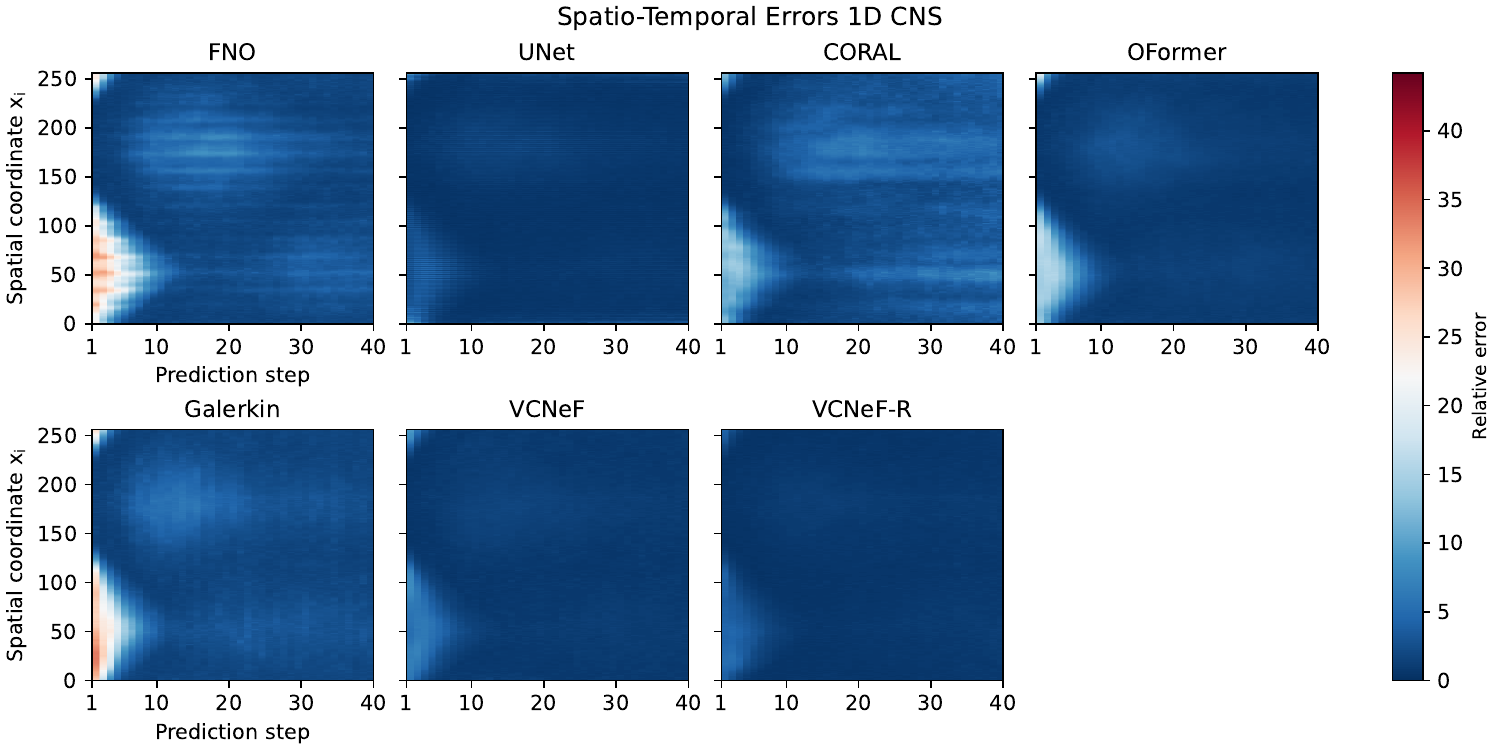}
        \caption{Error heatmap for 1D CNS with $\eta = \zeta = 0.007$. Models are trained and evaluated on spatial resolution $s = 256$ and temporal resolution $N_t = 41$. $t_0$ (not depicted above) is the initial condition.}
        \label{fig:cfd-heatmap}
    \end{center}
\end{figure}

Each of the operations in \cref{eq:rel-error-1d} is applied in a pointwise manner to the elements of $\boldsymbol{y_t}$ and $\boldsymbol{\hat{y_t}}$. We calculate the mean value over all test samples to get the final error heatmap. The error heatmap for 1D Burgers' equation in \cref{fig:burgers-heatmap} shows that all models, except for UNet and VCNeF, exhibit a very high error at the first and second timesteps (white area). Figures \ref{fig:advection-heatmap} and \ref{fig:cfd-heatmap} show that FNO and Galerkin Transformer have a very high error on a few spatial coordinates for 1D Advection and 1D CNS.

\FloatBarrier
\subsubsection{Temporal Error}
\begin{figure}[!htb]
    \begin{minipage}[b]{0.5\linewidth}
        \centering
        \includegraphics[width=1\textwidth]{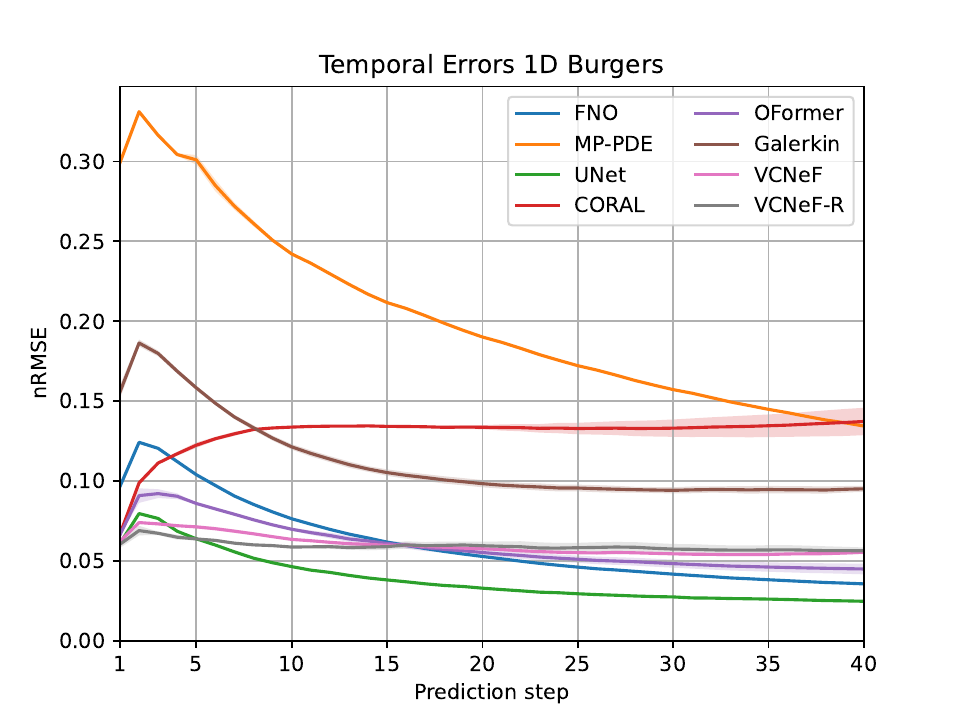}
        \label{subfig:temporal-error-burgers}
    \end{minipage}
    \hfill
    \begin{minipage}[b]{0.5\linewidth}
        \centering
        \includegraphics[width=1\textwidth]{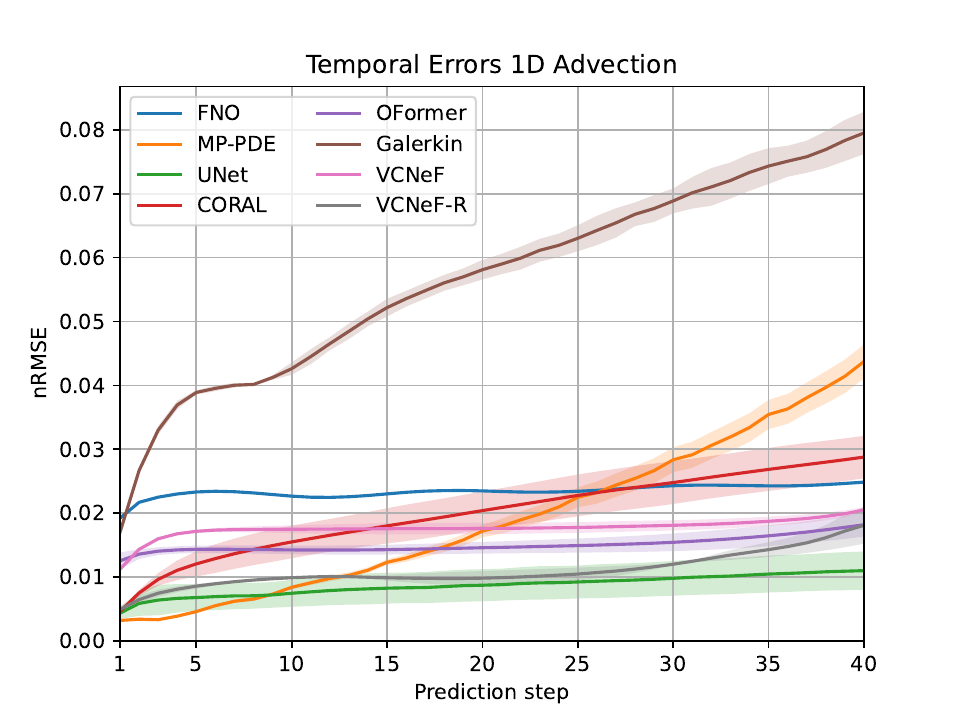}
        \label{subfig:temporal-error-advection}
    \end{minipage}
    \vspace{0.3em} 
    
    \centering
    \begin{minipage}{0.5\linewidth}
      \centering
      \includegraphics[width=1\textwidth]{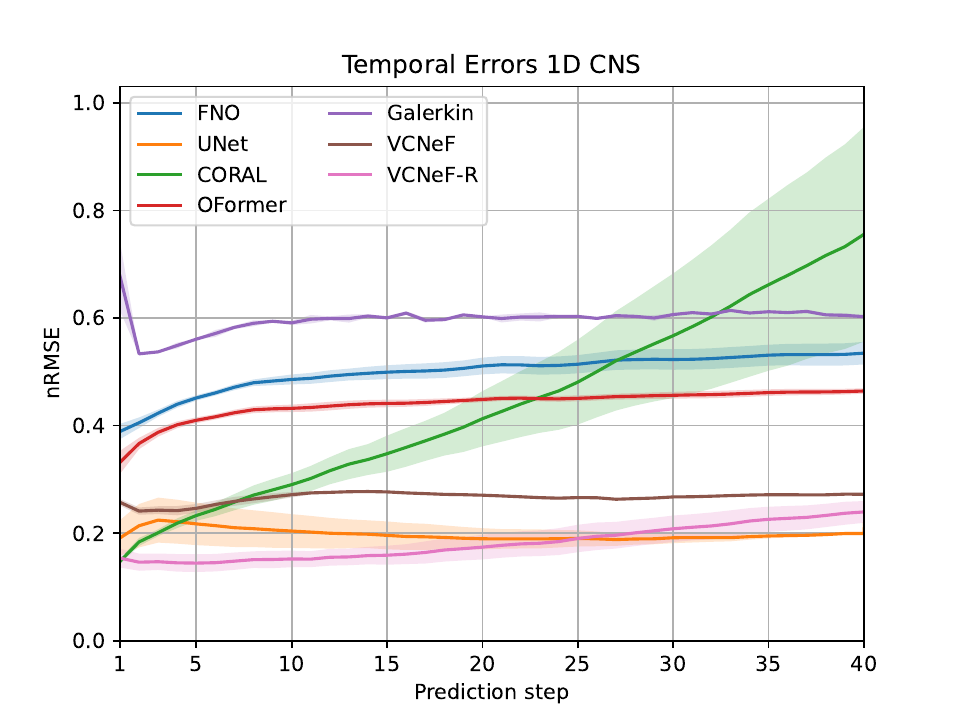}
      \label{subfig:temporal-error-1d-cns}
    \end{minipage}

  \caption{Temporal error of the models considered. The confidence band shows the standard deviation. $t_0$ (not visualized above) is the initial condition.}
  \label{fig:temporal_error}
\end{figure}

The following section shows the temporal error of the baselines and proposed model. Figure \ref{fig:temporal_error} shows the temporal error of the models for 1D Burgers, 1D Advection, and 1D CNS.

\FloatBarrier
\subsection{(Q2): Generalization to Unseen PDE Parameter Values}
\label{app:pde-param-generalization-results}
\begin{figure}[!htb]
    \begin{center}
        \includegraphics[width=0.99\textwidth]{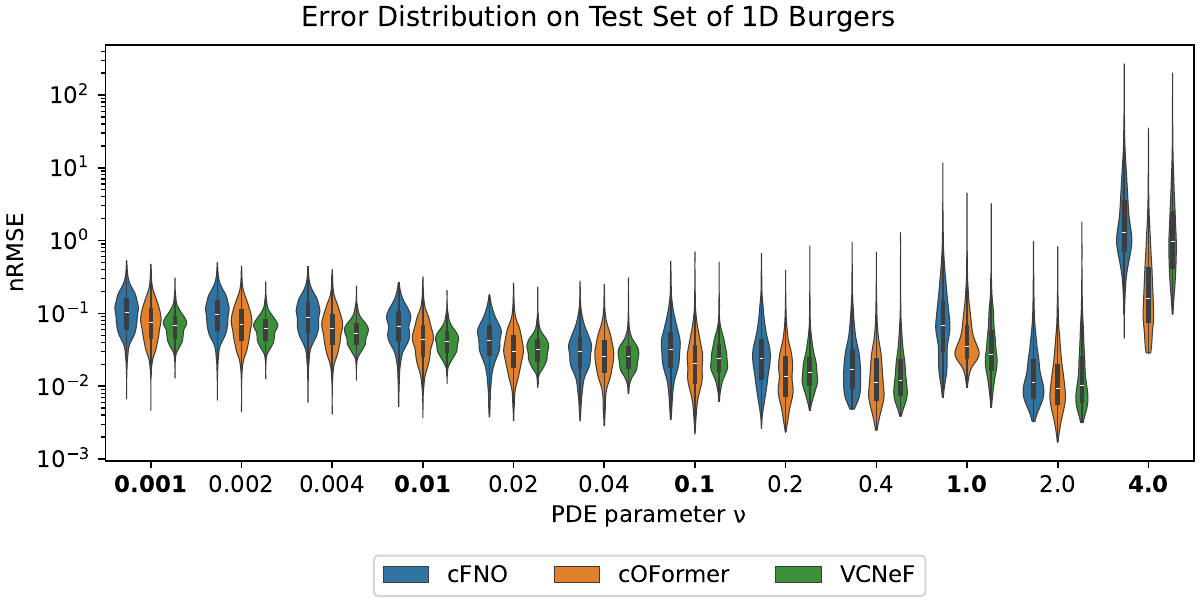}
        \caption{Error distribution of samples in the test set of 1D Burgers. Boldfaced are the unseen PDE parameter values.}
        \label{fig:violin-multi-burgers}
    \end{center}
\end{figure}

\begin{figure}[!htb]
    \begin{center}
        \includegraphics[width=0.99\textwidth]{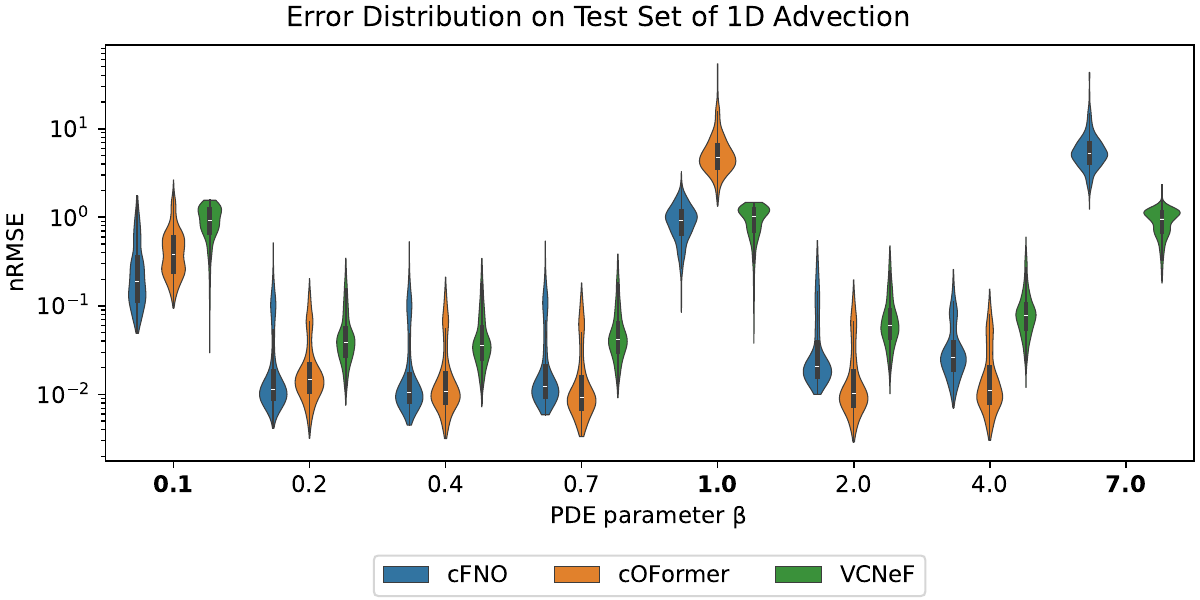}
        \caption{Error distribution of samples in the test set of 1D Advection. Boldfaced are the unseen PDE parameter values. Values for cOFormer and $\beta=7.0$ are missing since the model produced NaN at inference time.}
        \label{fig:violin-multi-advection}
    \end{center}
\end{figure}

\begin{figure}[!htb]
    \begin{center}
        \includegraphics[width=0.99\textwidth]
        {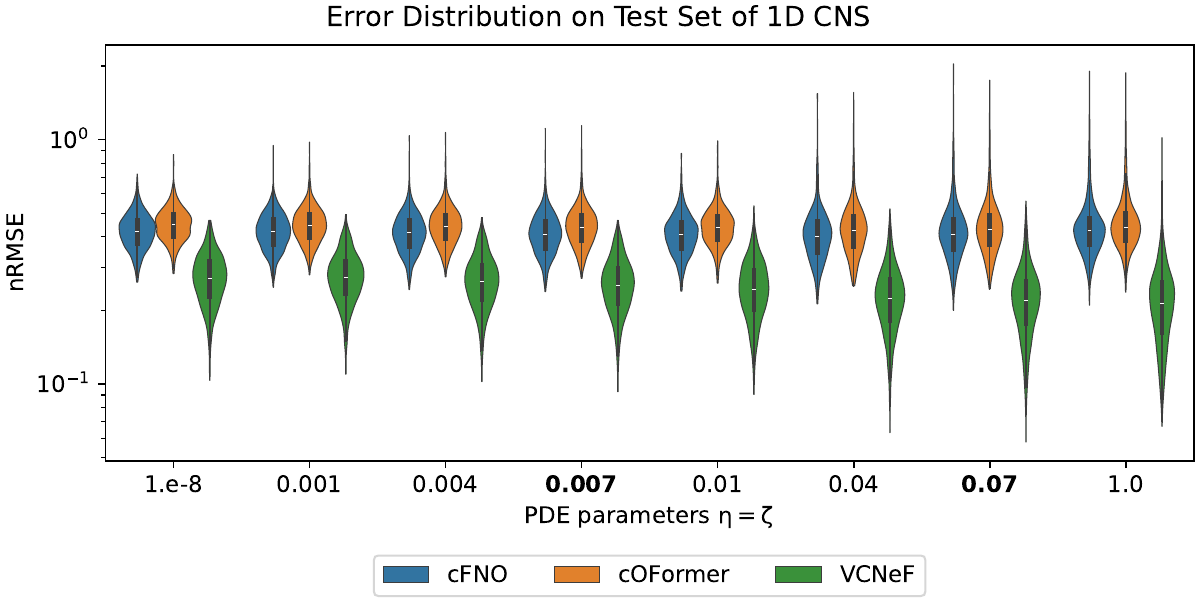}
        \caption{Error distribution of samples in the test set of 1D CNS. Boldfaced are the unseen PDE parameter values.}
        \label{fig:violin-multi-cfd-appendix}
    \end{center}
\end{figure}

We test VCNeF's generalization capabilities to unseen PDE parameter values by training it on a set of PDE parameter values and testing it on a different set of unseen PDE parameter values. We use cFNO \cite{cape-takamoto:2023} and cOFormer as the state-of-the-art baselines. Both models have been adapted to encode the PDE parameter as an additional input channel. Figures \ref{fig:violin-multi-burgers}, \ref{fig:violin-multi-advection}, \ref{fig:violin-multi-cfd-appendix} show the error distribution over the corresponding PDE parameter values and test sets.

\FloatBarrier
\subsection{(Q5): Inference Time and Memory Consumption}
\label{app:inference-times}
In traditional numerical solvers, the simulation time of trajectories of a given PDE is influenced by several factors, such as the value of the PDE parameter, efficiency of software implementation, the type and order of the numerical algorithm, discretization mesh, etc. On the contrary, the inference (simulation) time of ML models is agnostic to factors such as the PDE parameter value or order of numerical algorithm, which is one of the huge advantages of Neural PDE surrogates. \cref{tab:inference-times-verbose} demonstrates that the inference times of the proposed model scale better when compared to other transformer-based baselines. However, the speed-up results in a higher memory consumption. The model can also be used to do inference in a sequential fashion, which reduces memory consumption but increases the inference time. Nevertheless, it is still faster than OFormer, and the memory requirement remains the same even for extended rollout durations.

\begin{table}[!htb]
    \begin{center}
        \small
        \begin{tabular}{ clrlr }
            \toprule
            Prediction steps & Model & Inference time [ms] & & GPU memory consumption [MiB] \\
            \midrule
            \multirow{4}{*}{40} & FNO & 917.77 & $^{\pm2.51\ }$ & 716 \\
            & Galerkin & 2415.99&$^{\pm54.56}$ & 632 \\
            & OFormer & 6025.75&$^{\pm12.75}$ & 990 \\
            & VCNeF & 2244.04&$^{\pm6.65}$  & 4724 \\
            & VCNeF sequential & 4853.17&$^{\pm75.29}$  & 644 \\
            \midrule
            \multirow{4}{*}{80} & FNO & 1912.19&$^{\pm56.03}$ & 716 \\
            & Galerkin & 4940.80&$^{\pm89.44}$ & 632 \\
            & OFormer & 12081.98&$^{\pm19.39}$ & 990 \\
            & VCNeF & 4422.65&$^{\pm4.11}$  & 9284 \\
            & VCNeF sequential & 9701.80&$^{\pm84.48}$  & 644 \\
            \midrule
            \multirow{4}{*}{120} & FNO & 2808.04&$^{\pm82.22}$  & 716 \\
            & Galerkin & 7908.18&$^{\pm96.52}$ & 644 \\
            & OFormer & 17965.47&$^{\pm14.19}$ & 988 \\
            & VCNeF & 6606.41&$^{\pm3.00}$  & 13638 \\
            & VCNeF sequential & 14577.00&$^{\pm112.83}$  & 644 \\
            \midrule
            \multirow{4}{*}{160} & FNO & 3733.10&$^{\pm62.94}$ & 716  \\
            & Galerkin & 10295.78&$^{\pm116.50}$ & 644 \\
            & OFormer & 24108.24&$^{\pm6.45}$ & 990 \\
            & VCNeF & 6084.04&$^{\pm9.37}$  & 18871 \\
            & VCNeF sequential & 19449.80&$^{\pm113.73}$  & 644 \\
            \midrule
            \multirow{4}{*}{200} & FNO & 4614.21&$^{\pm97.52}$ & 718  \\
            & Galerkin & 13151.47&$^{\pm93.95}$ & 644 \\
            & OFormer & 29986.81&$^{\pm6.35}$ & 990 \\
            & VCNeF & 7584.48&$^{\pm1.86}$  & 22328 \\
            & VCNeF sequential & 24252.38&$^{\pm101.41}$  & 644 \\
            \midrule
            \multirow{4}{*}{240} & FNO & 5572.07&$^{\pm109.23}$ & 716  \\
            & Galerkin & 15600.60&$^{\pm262.51}$ & 644 \\
            & OFormer & 35900.51&$^{\pm6.71}$ & 988 \\
            & VCNeF & 8935.28&$^{\pm7.08}$  & 26662 \\
            & VCNeF sequential & 29063.89&$^{\pm79.58}$  & 668 \\
            \bottomrule
        \end{tabular}
        \caption{Inference times and GPU memory consumptions of different models trained and evaluated on the 1D Burgers' equation with a spatial resolution of 256, predicting different numbers of timesteps in the future. ``VCNeF sequential'' means a VCNeF model that computes the solutions for all timesteps sequentially.}
        \label{tab:inference-times-verbose}
    \end{center}
\end{table}

\FloatBarrier
\section{Qualitative Results}

Here, we compare visualizations of the predictions vs ground truth for 1D Advection, Burgers, and 2D Compressible Navier-Stokes PDEs. The 2D CNS dataset has four channels, namely density velocity-x and velocity-y, and pressure, and we visualize the predictions of our VCNeF model with the ground truth data.

\begin{figure}[!htb]
    \begin{center}
        \includegraphics[width=0.8\textwidth]{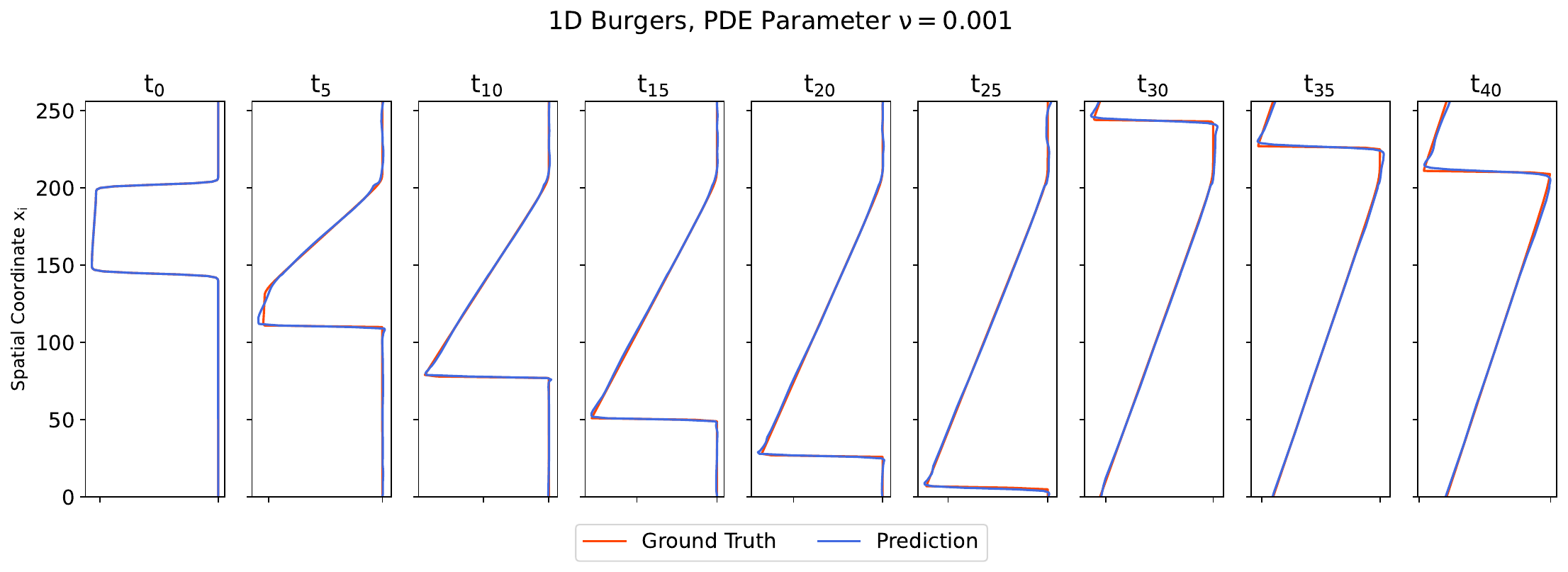}
        \caption{Example prediction's of VCNeF for 1D Burgers with $N_t = 41$.}
        \label{fig:example-burgers-1}
    \end{center}
\end{figure}

\begin{figure}[!htb]
    \begin{center}
        \includegraphics[width=0.8\textwidth]{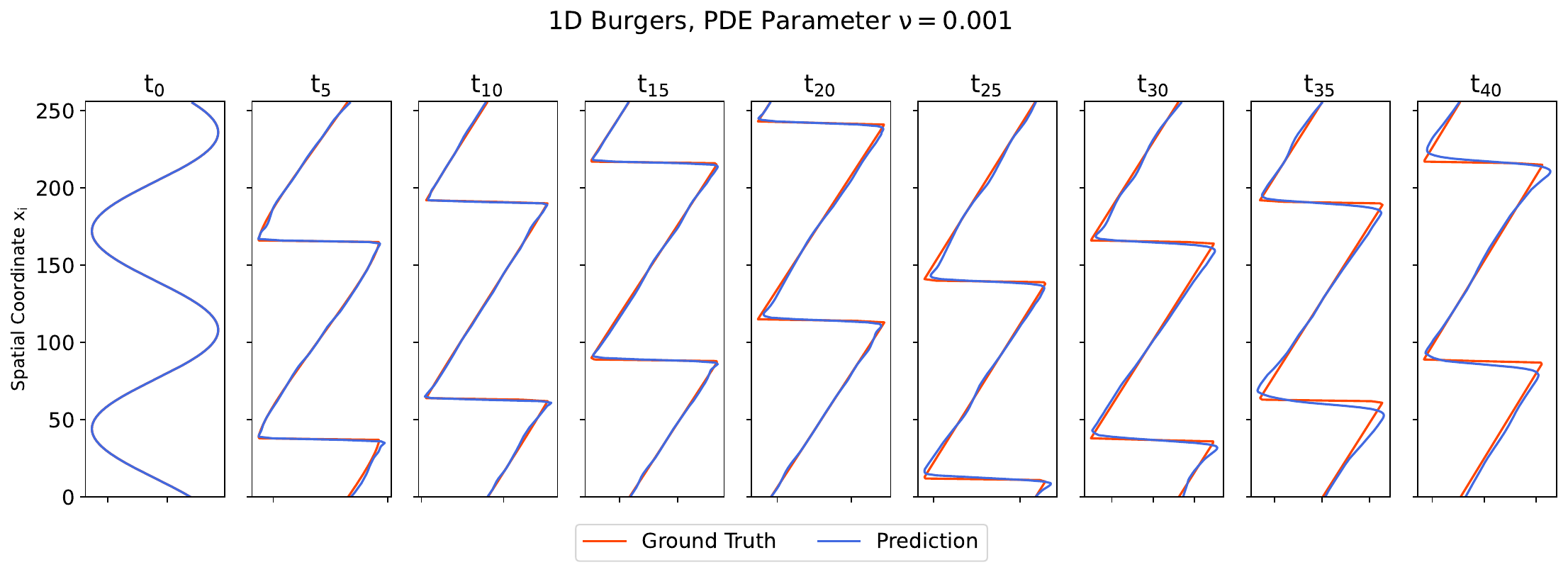}
        \caption{Example prediction of VCNeF for 1D Burgers with $N_t = 41$.}
        \label{fig:example-burgers-2}
    \end{center}
\end{figure}

\begin{figure}[!htb]
    \begin{center}
        \includegraphics[width=0.8\textwidth]{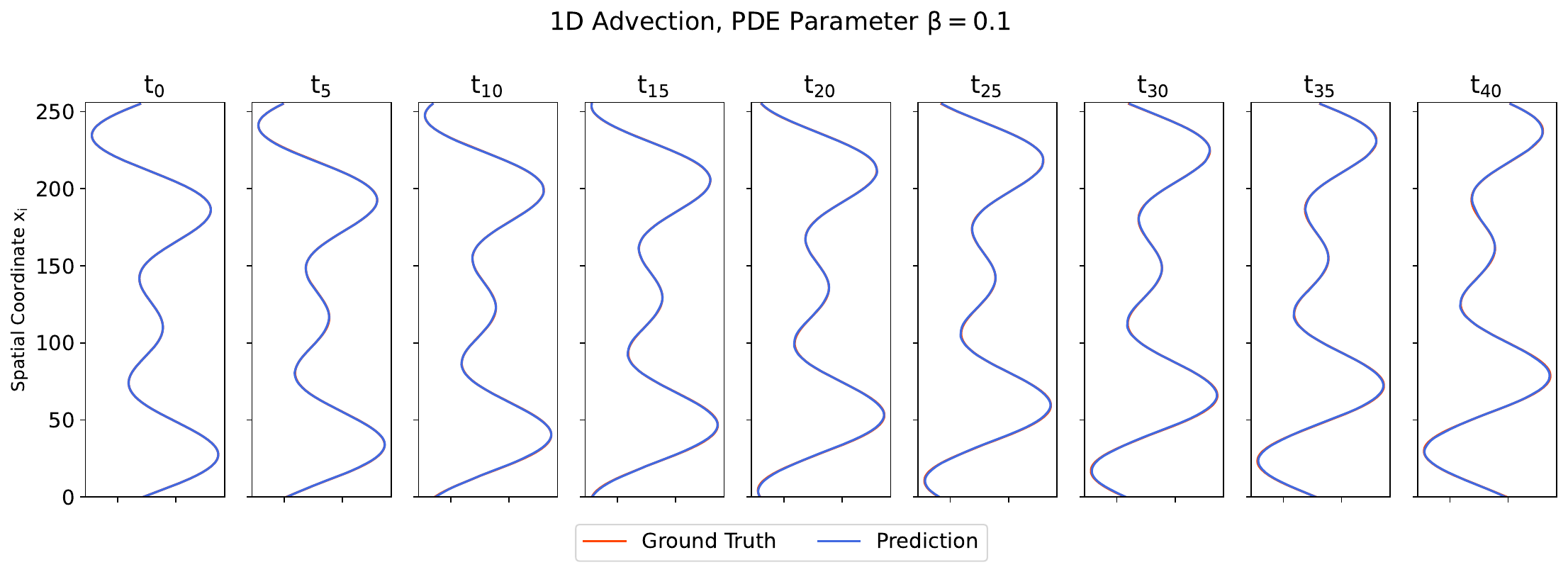}
        \caption{Example prediction of VCNeF for 1D Advection with $N_t = 41$.}
        \label{fig:example-advection-1}
    \end{center}
\end{figure}

\begin{figure}[!htb]
    \begin{center}
        \includegraphics[width=0.8\textwidth]{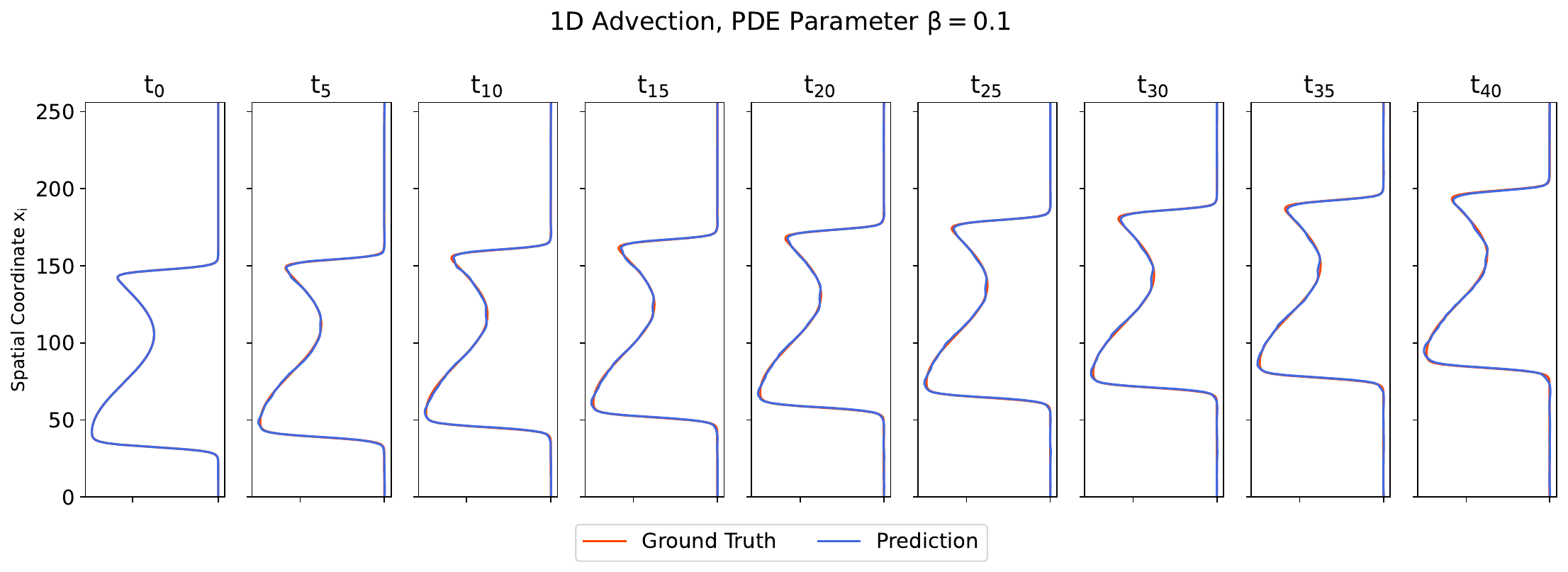}
        \caption{Example prediction of VCNeF for 1D Advection with $N_t = 41$.}
        \label{fig:example-advection-2}
    \end{center}
\end{figure}

\begin{figure}[!htb]
    \begin{center}
        \includegraphics[width=0.75\textwidth]{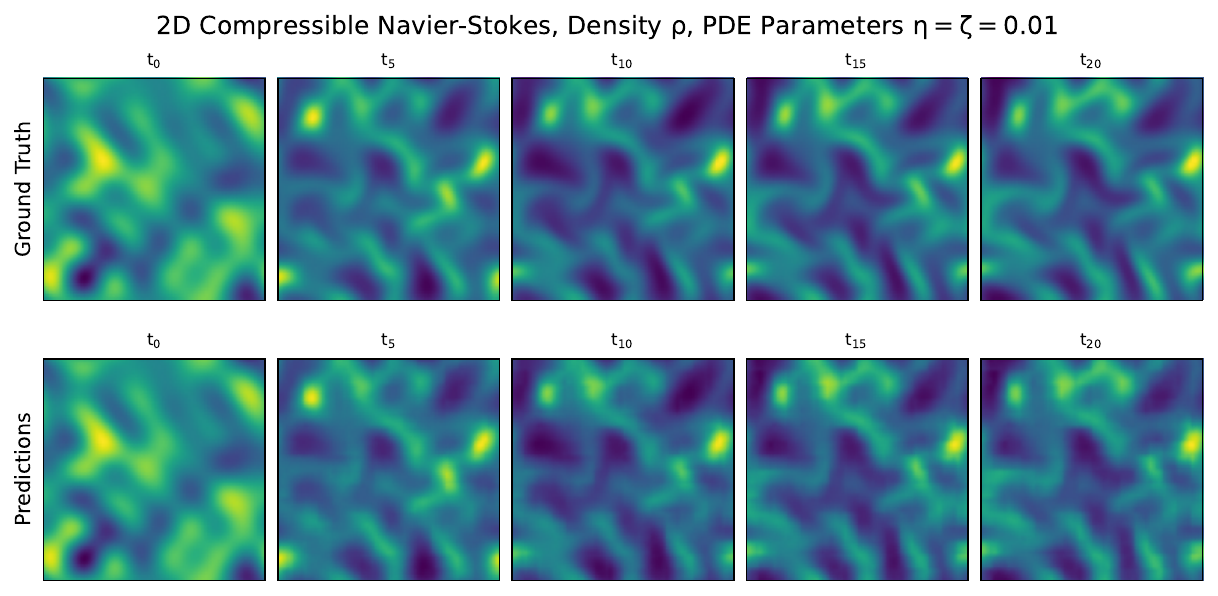}
        \caption{Example prediction of VCNeF for the density channel of 2D compressible Navier-Stokes with $N_t = 21$.}
        \label{fig:density-2d-cfd}
    \end{center}
\end{figure}

\begin{figure}[!htb]
    \begin{center}
        \includegraphics[width=0.75\textwidth]{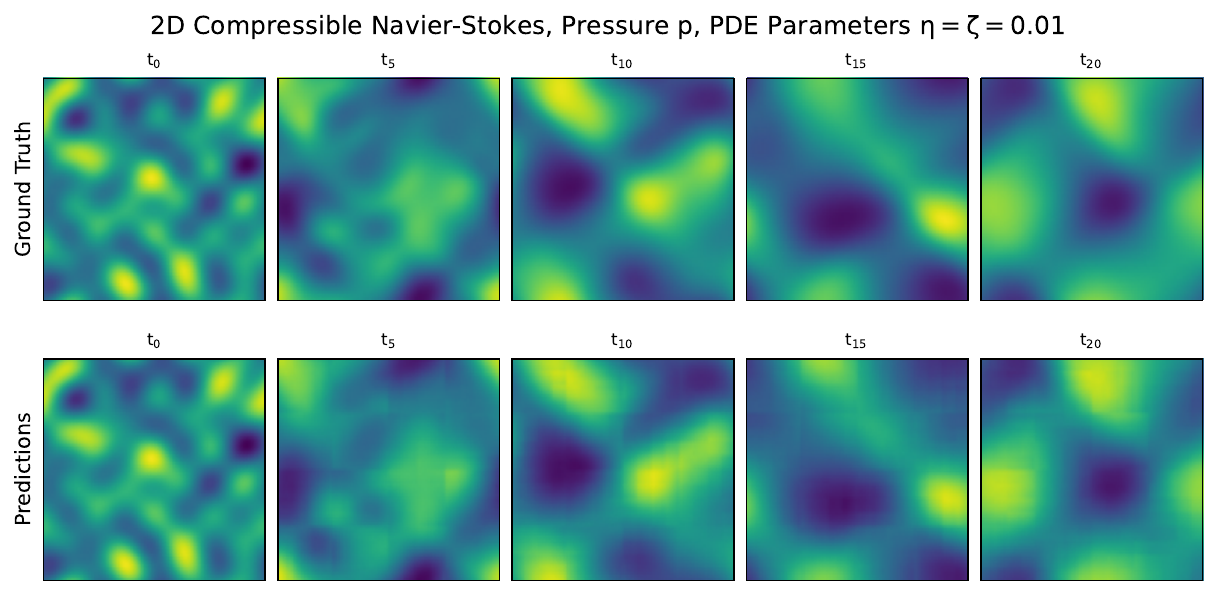}
        \caption{Example prediction of VCNeF for the pressure channel of 2D compressible Navier-Stokes with $N_t = 21$.}
        \label{fig:pressure-2d-cfd}
    \end{center}
\end{figure}

\begin{figure}[!htb] 
    \begin{center}
        \includegraphics[width=0.75\textwidth]{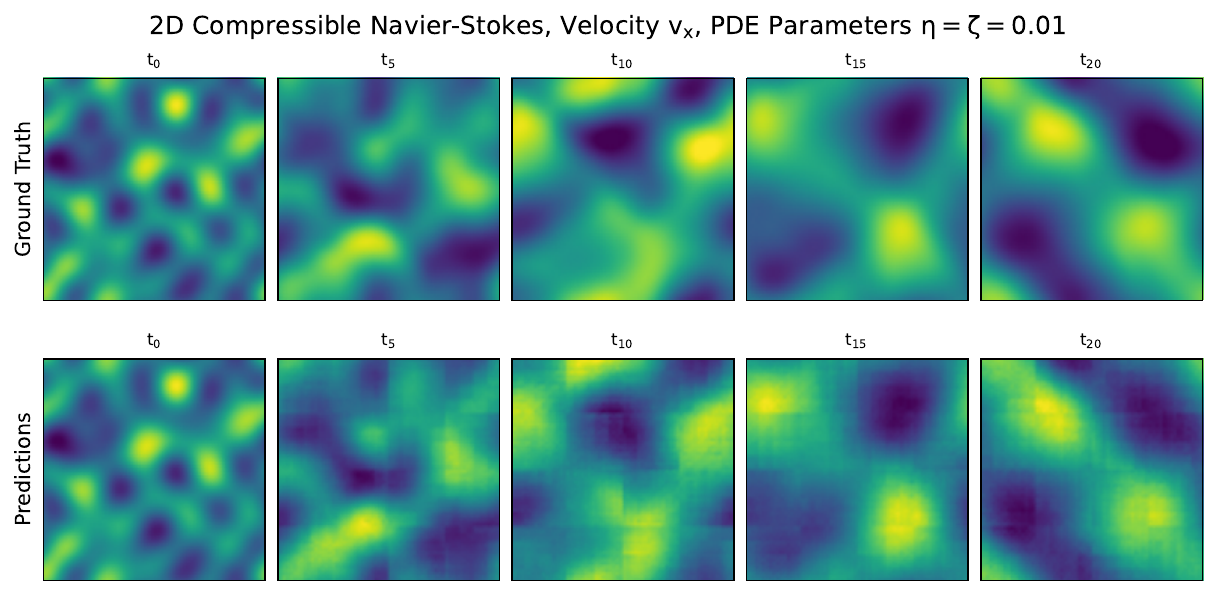}
        \caption{Example prediction of VCNeF for the velocity x-axis channel of 2D compressible Navier-Stokes with $N_t = 21$.}
        \label{fig:vx-2d-cfd}
    \end{center}
\end{figure}

\end{document}